%% file: main_arxiv.tex
\RequirePackage[OT1]{fontenc}
\documentclass[[10pt,twocolumn,twoside]{IEEEtran}

\usepackage{times}  %
\usepackage{helvet}  %
\usepackage{courier}  %
\usepackage{url}  %
\usepackage{graphicx}  %
\usepackage{color}
\usepackage{amsmath}
\usepackage{amssymb}
\usepackage{multirow}
\usepackage{multicol}

\makeatletter
\let\MYcaption\@makecaption
\makeatother
\usepackage[font=footnotesize]{subcaption}
\makeatletter
\let\@makecaption\MYcaption
\makeatother

\usepackage{balance}
\usepackage{hyperref}
\usepackage{epsfig}
\usepackage{tabularx}
\usepackage{booktabs}
\usepackage{tabulary}
\usepackage{xspace}
\usepackage{enumitem}
\usepackage[table,dvipsnames]{xcolor}
\usepackage{lipsum}
\usepackage{tikz}
\usepackage{tabularx}
\usepackage[ruled, linesnumbered]{algorithm2e}
\usepackage{algpseudocode}  
\usepackage{colortbl}
\usepackage{enumitem}
\usepackage{soul}
\usepackage{xspace}
\usepackage{multirow}
\usepackage{mathtools}
\usepackage{rotating}
\usepackage{url}
\usepackage{array}
\usepackage[export]{adjustbox}
\usepackage{ctable}
\usepackage{scalerel}
\usepackage{lineno}
\usepackage{lipsum}
\usepackage{dblfloatfix}

\definecolor{lime}{HTML}{A6CE39}
\DeclareRobustCommand{\orcidicon}{%
	\begin{tikzpicture}
	\draw[lime, fill=lime] (0,0) 
	circle [radius=0.16] 
	node[white] {{\fontfamily{qag}\selectfont \tiny ID}};
	\draw[white, fill=white] (-0.0625,0.095) 
	circle [radius=0.007];
	\end{tikzpicture}
	\hspace{-3mm}
}

\foreach \x in {A, ..., Z}{%
	\expandafter\xdef\csname orcid\x\endcsname{\noexpand\href{https://orcid.org/\csname orcidauthor\x\endcsname}{\noexpand\orcidicon}}
}

\hypersetup{
    colorlinks=true,
    urlcolor=blue,
}
\def\best#1{\textbf{#1}}

\def\etc{%
    \@ifnextchar{.}%
        \emph{etc}%
        \emph{etc.\@\xspace}%
}

\newcommand{\secref}[1]{\mbox{Section~\ref{#1}}}

\newcommand{\figref}[1]{\mbox{Figure~\ref{#1}}}

\renewcommand{\eqref}[1]{\mbox{Equation~(\ref{#1})}}

\newcommand{\tabref}[1]{\mbox{Table~\ref{#1}}}

\newcolumntype{L}[1]{>{\raggedright\arraybackslash}p{#1}}
\newcolumntype{C}[1]{>{\centering\arraybackslash}p{#1}}
\newcolumntype{R}[1]{>{\raggedleft\arraybackslash}p{#1}}

\newcommand{\ie}{{i.e.}\@\xspace}
\newcommand{\eg}{{e.g.}\@\xspace}
\newcommand{\etal}{{\it et~al.}\@\xspace}

\def\Vec#1{{\boldsymbol{#1}}}
\def\Mat#1{{\boldsymbol{#1}}}
\def\Set#1{{\mathcal{#1}}}
\def\tVec#1{{\tilde{\boldsymbol{#1}}}}

\newcommand{\Norm}[1]{\left\lVert#1\right\rVert}

\newcommand{\PrimConcept}[1]{$\mathsf{#1}$}
\newcommand{\CplxConcept}[2]{\mbox{$\langle \mathsf{#1}, \mathsf{#2} \rangle$}}
\newcommand{\MErr}[2]{${#1}{\scriptscriptstyle\pm{#2}}$}

\DeclareMathOperator*{\argmax}{\arg\max}

\graphicspath{{./figs/}}

\newcommand{\revcolorB}{black}
\newcommand{\revtextB}[1]{\textcolor{\revcolorB}{#1}}

\begin{document}
\title{Relation-aware Compositional Zero-shot Learning for Attribute-Object Pair Recognition}

\author{
    Ziwei~Xu\orcidA,~\IEEEmembership{Student~Member,~IEEE,}
    Guangzhi~Wang\orcidB{},
	Yongkang~Wong\orcidC{},~\IEEEmembership{Member,~IEEE,}
	Mohan~Kankanhalli\orcidD{},~\IEEEmembership{Fellow,~IEEE}
	\IEEEcompsocitemizethanks{
        © 2021 IEEE.\@ Personal use of this material is permitted.  Permission from IEEE must be obtained for all other uses, in any current or future media, including reprinting/republishing this material for advertising or promotional purposes, creating new collective works, for resale or redistribution to servers or lists, or reuse of any copyrighted component of this work in other works.

		Ziwei~Xu, Yongkang~Wong and Mohan~Kankanhalli are with the School of Computing, 
		National University of Singapore
		(email: \{ziwei-xu, wongyk, mohan\}@comp.nus.edu.sg).
		Guangzhi~Wang is with NUS Graduate School for Integrative Sciences \& Engineering 
		(email: guangzhi.wang@u.nus.edu).
        The corresponding author is Ziwei~Xu.
	}
}

\markboth{Xu~\MakeLowercase{\etal}: Relation-aware Compositional Zero-shot Learning for Attribute-Object Pair Recognition}%
{Xu~\MakeLowercase{\etal}: Relation-aware Compositional Zero-shot Learning for Attribute-Object Pair Recognition}

\maketitle

\input{sec_abstract.tex}


\IEEEpeerreviewmaketitle

\input{sec_introduction}

\input{sec_related_work}
\input{sec_methodology}

\input{sec_experiment}

\input{sec_conclusion}

\nolinenumbers

\ifCLASSOPTIONcaptionsoff
  \newpage
\fi

\bibliographystyle{IEEEtran}
\bibliography{reference}

\input{sec_biography}

\end{document}

%% file: sec_abstract.tex
\begin{abstract}

This paper proposes a novel model for recognizing images with composite attribute-object concepts,
notably for composite concepts that are unseen during model training.
We aim to explore the three key properties required by the task --- relation-aware, consistent, and decoupled --- to learn rich and robust features for primitive concepts that compose attribute-object pairs.
To this end, we propose the Blocked Message Passing Network (BMP-Net).
The model consists of two modules. 
The concept module generates semantically meaningful features for primitive concepts, 
whereas the visual module extracts visual features for attributes and objects from input images.
A message passing mechanism is used in the concept module to capture the relations between primitive concepts.
Furthermore, to prevent the model from being biased towards seen composite concepts and reduce the entanglement between attributes and objects, we propose a blocking mechanism that equalizes the information available to the model for both seen and unseen concepts.
Extensive experiments and ablation studies on two benchmarks show the efficacy of the proposed model.

\end{abstract}

%% file: sec_introduction.tex
\section{Introduction}
\label{sec:introduction}

In visual recognition tasks, a detectable entity can often be described as a composition of primitive concepts that characterize different appearance cues. 
For example, a {\CplxConcept{red}{tomato}} or a {\CplxConcept{tall}{person}}, where \PrimConcept{red}, \PrimConcept{tall}, \PrimConcept{tomato} and \PrimConcept{person} are primitive concepts.
In other words, most of the entities contain \textit{composite} concepts. 
However, given the variety of primitive concept types and a large number of concepts in each of those types, it is computationally intractable to enumerate all the possible combinations of primitive concepts and model them one-by-one. 
It is, however, more tractable to model primitive concepts separately in both concept feature space and visual feature space and then match concepts with their visual correspondences for recognition.
In this way, the complexity required to describe the composite concepts drops from {$O(N^2)$} to {$O(N)$} if we have $N$ primitive concepts to consider. 

\input{depd/fig-teaser}

The separate modeling of primitive concepts %
gives rise to an interesting observation: since {$ N^2 \texttt{>>} N $} when {$ N \texttt{>>} 1 $}, we are gaining knowledge about a much larger set of unseen composite concepts from a small set of seen composite concepts.
Therefore, this task is also referred to as \emph{compositional zero-shot learning} (CZSL).
This is analogous to human learning~\cite{Duncan2017ComplexityAC}.
Assume we only have seen images of {\CplxConcept{dry}{dog}} and {\CplxConcept{wet}{cat}}, one can still easily imagine a {\CplxConcept{wet}{dog}} and recognize an entity of that concept in images 
(as illustrated in \figref{fig:teaser}). 
The same task, however, is non-trivial for a computational model.

The challenge of CZSL is three-fold.
First, different primitive concepts relate to each other in different ways.
For example, \PrimConcept{sharp} is the antonym of \PrimConcept{blunt}, \PrimConcept{red} and \PrimConcept{yellow} reflect specific colors, and \PrimConcept{puppy} and \PrimConcept{dog} refer to different ages of the same species.
Modeling each primitive concept regardless of others could result in a loss of important information in resulting features.
This calls for a \emph{relation-aware} model.
Second, since the model only has access to seen composite concepts during training, the learning could excessively rely on a limited number of relations between primitive concept pairs and downplay the importance of others (\eg~unobserved relations between unseen primitive concept pairs).
To balance this effect, the training and inference process should be \emph{consistent} for both seen and unseen composite concepts.
Third, composite concepts tend to get entangled in the real world.
For example, \PrimConcept{tomato} appears frequently with \PrimConcept{red} or \PrimConcept{green} but rarely with \PrimConcept{blue}.
This bias results in the model's high dependency on objects when inferring attributes, and vice versa, which suggests that the model should deal with attributes and objects in a \emph{decoupled} way.
All three challenges could severely affect the model's ability to recognize unseen composite concepts.
Prior works~\cite{Misra2017FromRW, Xian2017FeatureGN, Nagarajan2018AttributesAO, Purushwalkam2019TaskDrivenMN, Wei2019AdversarialFC} on CZSL mainly focus on resolving the second challenge, while limited attempts have been made towards resolving the first and third challenge by introducing loss terms related to linguistic cues~\cite{Nagarajan2018AttributesAO} or fine-grained structures~\cite{Wei2019AdversarialFC}.

In this paper, we aim to recognize entities of attribute-object composite concepts in images by overcoming the aforementioned challenges.
Towards this goal, we propose the Blocked Message Passing Network (BMP-Net).
The network consists of a concept module and a visual module.
The concept module takes attribute/object concepts as input and generates the corresponding concept features, while the visual module extracts the attribute/object visual features from an input image.
The composite concept in the image is recognized by matching attribute and object features from both modules.
To make the model \emph{relation-aware}, we propose a concept module that generates concept features by message passing (MP) between primitive concepts in a graph structure.
The relations between concepts are captured by a key-query based attention mechanism, which determines the strength of edges during MP.
To build a \emph{consistent} and \emph{decoupled} model, we introduce a blocking mechanism to our model to reduce the unnecessary and bias-prone information flow.
The blocking is used in two ways.
First, towards a consistent model, we block edges between selected concepts during MP to reduce the model's reliance on the information from seen pairs during training and inference.
Second, towards a decoupled model, we randomly block either the attribute or the object branch during training to reduce the co-occurrence of attributes and objects.

The key contributions of this work are:
(1)~We propose a relation-aware concept module where the primitive concept features are generated from messages passed from other primitive concepts using a key-query based attention mechanism. 
By using this mechanism, our model can exploit the relations between different primitive concepts and generate rich concept features for effective recognition.
(2)~To reduce the bias towards seen composite concepts and the entanglement between attributes and objects, we propose the Blocked Message Passing (BMP) method, which includes edge blocking for the concept module and branch blocking during training.
This provides a remarkable improvement in recognizing unseen composite concepts without sacrificing the performance on seen concepts.
(3)~We achieve state-of-the-art performance on UT-Zappos~\cite{Yu2014FineV} and MIT-States~\cite{Isola2015DS}.
Extensive ablation studies examine the impact of various components of our method.

%% file: depd/fig-teaser.tex
\begin{figure}
	\centering
	\includegraphics[width=0.95\columnwidth]{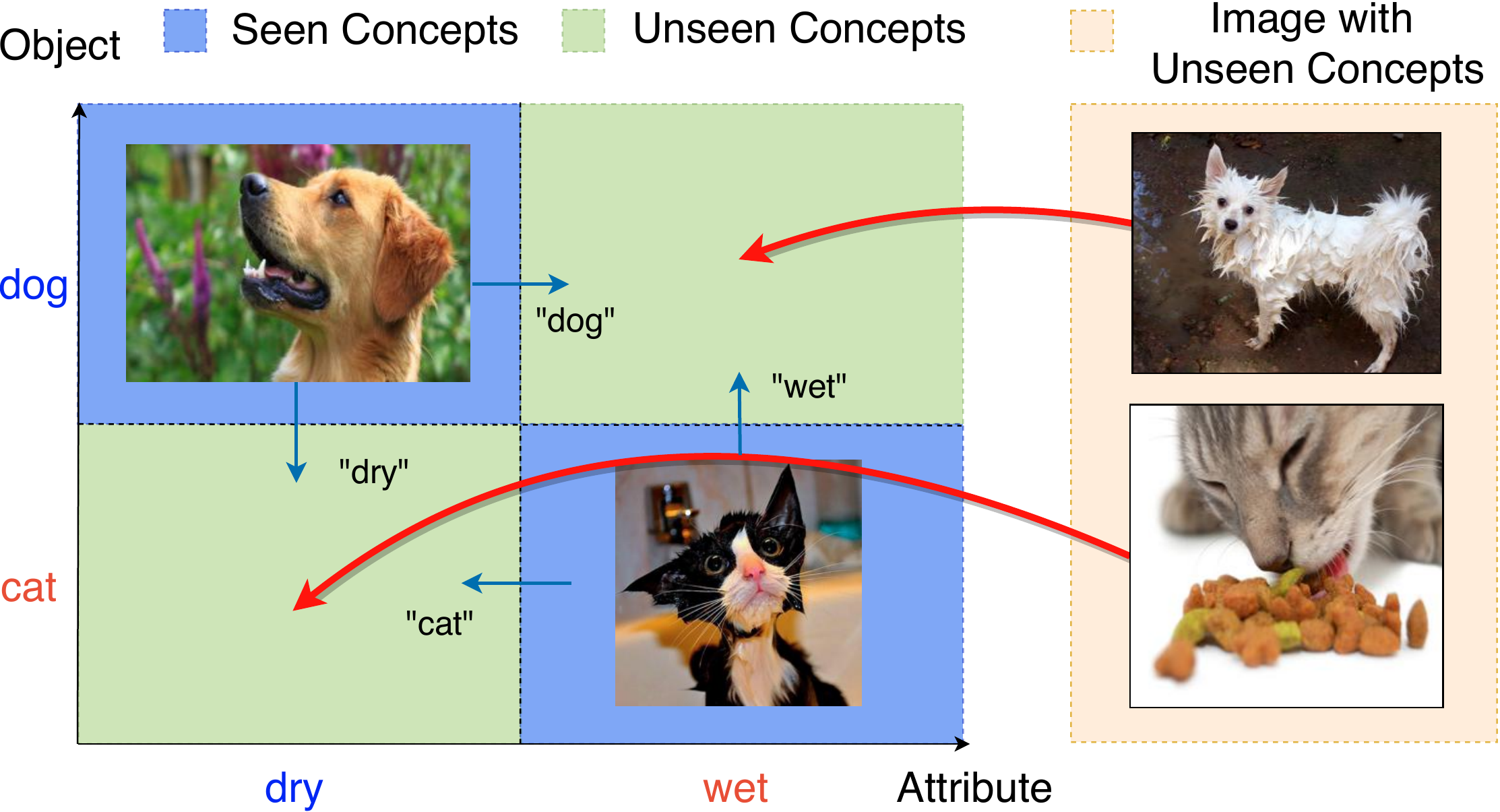}
    \vspace{-1em}
	\caption{\label{fig:teaser} 
	    Given training images of {\small \CplxConcept{\textcolor{blue}{dry}}{\textcolor{blue}{dog}}} and {\small \CplxConcept{\textcolor{red}{wet}}{\textcolor{red}{cat}}} concepts, can a model gain any knowledge about unseen concepts (\ie~{\small \CplxConcept{\textcolor{blue}{dry}}{\textcolor{red}{cat}}} and {\small \CplxConcept{\textcolor{red}{wet}}{\textcolor{blue}{dog}}})? 
	    An intuitive solution is to model the primitive concepts separately, and then match them with visual features extracted from images.}
\end{figure}

%% file: sec_related_work.tex
\section{Related Work}\label{sec:related_work}

\subsection{Attribute and its Composition}

Attributes have been widely studied in the literature.
In the vision community, attributes are considered as mid-level semantics invariant across different objects or domains~\cite{Hwang2011SharingFB,Gan2016LearningAE}.
The detected attributes can benefit downstream tasks in multiple ways.
For example, it can serve as node features in graph similarity learning~\cite{Chang2014FactorizedSLN} for multimodal content retrieval~\cite{Zhang2020JointAttributM}.
Wang~\etal~\cite{Wang2012RecommendingFG} propose a topic model using attributes detected in Flickr images for group recommendation.
Visual attributes are also widely used in 
zero/few-shot image classification~\cite{Akata2013LabelEmbeddingFA, Deng2014LargeScaleOC, Farhadi2009DescribeO, Parikh2011RelAttr, Rahman2018UnifiedAC}, 
action recognition~\cite{Liu_CVPR_2011,Zhang_JEET_2015}, 
and image captioning~\cite{Kulkarni2013BabyTalk}.
These methods consider visual attributes as categories and train discriminative models to recognize them from images or videos.
Attributes are also modeled as state transformations, which is beneficial for modeling visual appearances and physical state changes of objects in images~\cite{Nagarajan2018AttributesAO,Isola2015DS}. 
Moreover, it can be used to jointly model manipulative actions and object states for action recognition~\cite{Alayrac2017JointD,Liu2017JR,Zhuo_ACMMM_2019}.

The composition of visual attributes can be formulated as adjective-object (\eg~\PrimConcept{red}-\PrimConcept{wine}), verb-object (\eg~\PrimConcept{read}-\PrimConcept{book}), or subject-verb-object (\eg~\PrimConcept{kid}-\PrimConcept{read}-\PrimConcept{book}).
The latter two are generally referred to as visual relation or human-object interaction~\cite{Xu2020InteractYouIntend}.
Early works in this field use probabilistic models~\cite{Gupta2007ObjectAction} and local context like spatial configuration~\cite{Yao2010Grouplet,Sadeghi2011RecognitionVP} to decode complex attributes from visual input.
Recently, Graph Neural Networks (GNN) have gained popularity as a powerful relation model for this task~\cite{Zhu2018LearningHOI,Xu2019LearningDetectHOI,Xie2019EmbeddingSK,Zhou2021CascadedParsingHOI}.
Our work belongs to the adjective-object category, which is detailed in \secref{ssec:czsl}.

\subsection{Zero-shot Learning (ZSL)}
This task aims at recognizing unseen categories with the help of seen categories and their semantic description~\cite{Lampert2009LearningTD, Changpinyo2016SynthesizedC, AlHalah2016RecoveringTM, Min2019DomainSpecificEN, Liu2019Attribute, Xian2019fvaegand2, Xie2020Region, Guo2020Novel, Shen2020Invertible, Gao2020CIGNN}.
In the traditional setting, the model is trained on the seen categories' images and evaluated on the images of unseen ones.
This setting is limited because of its strong assumption on the categories during evaluation~\cite{Xian2019ZSLComprehensive}.
In the generalized ZSL setting~\cite{Bendale2016TowardsOS, Chao2016AnES, Xian2019ZSLComprehensive, Chen2020Generalized}, the trained model is evaluated on the images of both seen and unseen categories.
Learning to align image features with semantic descriptions is crucial in ZSL.
Qi~\etal~\cite{Qi2017JointInterIntramodel} propose to learn transfer functions which propagates semantics between text and visual features.
Wang~\etal~\cite{Wang2018ZeroRSEKG} use a graph convolutional network to propagate semantic information between categories.
Chen~\etal~\cite{Chen2018ZeroVR} argue that non-discriminant semantic information for seen categories is lost during training and proposes an adversarial training method to preserve it.
Recently, ZSL employs the attention mechanism to align fine-grained local information with semantic descriptions. 
S$^2$GA~\cite{Yu2018StackedSA} introduces an attention model which assigns weights to local features based on their relevance to different semantic descriptions.
DAZLE~\cite{Huynh2020FineGZSL} employs a dense attribute-based mechanism to capture the fine-grained discriminative regions.
Apart from image recognition, zero-shot action and video analysis have also received attention for the scarceness of annotation.
For example, word vectors~\cite{Gan2015ExploringSIR}, external ontologies~\cite{Gan2016RecognizingAU}, and knowledge graphs~\cite{Gao2019IKnowRelationships} have been used to build links that transfer semantics from seen action classes to unseen ones.

\subsection{Compositional Zero-shot Learning (CZSL)}
\label{ssec:czsl}

The CZSL task, first introduced in~\cite{Misra2017FromRW}, requires a model to recognize a set of unseen \textit{composite} concepts with the help of a set of seen ones.
Different from conventional zero-shot learning
where a category is usually represented by a set of underlying \textit{seen} semantic descriptions,
the CZSL task does not assume any prior knowledge about primitive concepts.
Instead, it requires the model to learn the semantic representation for each primitive concept.
Previous works can be categorized into (1)~classifier composition, (2)~attribute-as-transformation, and (3)~feature synthesis.
The classifier composition methods merge classifiers for primitive concepts into classifiers for composite concepts, which can be done via linear transformations~\cite{Misra2017FromRW}, Boolean algebra~\cite{Cruz2018NeuralAO}, or tensor completion~\cite{Chen2014InferringAA}. 
Purushwalkam~\etal~\cite{Purushwalkam2019TaskDrivenMN} propose to express the recognition process as a combination of sub-tasks and train a set of small networks to solve them.
The drawback of classifier composition is that the bias is still present in the respective classifiers and accumulates as the classifiers are composed, which makes it less \textit{consisitent}.
The attribute-as-transformation methods model objects as vectors and attributes as transformations in the vector space. 
Nagarajan~\etal~\cite{Nagarajan2018AttributesAO} generate the representations of unseen attribute-object pairs by applying the corresponding attribute transformation matrix on the target object's vector. 
SymNet~\cite{Li2020SymmetryGA} exploits the symmetry principle to improve the modeling of the transformation process.
Compared with classifier composition, this formulation allows explicit regularization on attributes but disregards the semantic relationships between primitive concepts and is therefore not \textit{relation-aware}.
The feature synthesis methods directly model attributes, objects, and attribute-object pairs as vectors.
In~\cite{Kodirov2017SemanticAF, Nan2019RecognizingUA}, primitive concepts are first identified and then merged to form features for unseen composite concepts.
Wei~\etal~\cite{Wei2019AdversarialFC} generate features for composite concepts using generative adversarial networks and proposed a quintuplet loss for fine-grained learning.
More recently, a causal feature synthesis method is proposed~\cite{Atzmon2020CausalCZR}.
The shared feature space allows for flexible constraints and relation modeling between any primitive concepts.
Our method belongs to the feature synthesis category. However, different from prior works, we achieve \textit{relation-aware}, \textit{consistent}, and \textit{decoupled} properties in one model by utilizing the relations between primitive concepts using the proposed BMP method.

\input{depd/fig-framework-overview}

\subsection{Attention Mechanism}
The attention mechanism was originally proposed by the NLP community to capture the dependencies between tokens in sequential data.
Its early applications include machine translation~\cite{Bahdanau2014NeuralMT}, where the attention is introduced between tokens of the input and output sequences.
The Transformer~\cite{Vaswani2017AttentionIA} have generalized the concept of attention
and has been proven useful in many other tasks like document understanding~\cite{Parikh2016ADA} and linguistic model pretraining~\cite{Radford2018ImprovingLU, Devlin2019BERTPO}.
It is also widely used in visual tasks like object segmentation~\cite{Wang2019ZSLVideoObject}, motion analysis~\cite{Shu2021SpatialtemporalCRN}, visual dialog~{\cite{Zhu2019ReasoningVD}}, and visual question answering~\cite{Liu2019ErasingbasedAL}.

The key-query attention mechanism used in this paper is inspired by the key-value memory networks~\cite{Miller2016KeyValueMN}.
In~\cite{Miller2016KeyValueMN}, a memory network~\cite{Sukhbaatar2015EndToEndMN} with a key-based addressing mechanism is proposed to solve the question answering problem.
The keys are used to calculate the attention score of input linguistic cues over all possible memory blocks and then used to retrieve information from target memory blocks as the answer.
Despite the success in various NLP tasks, to the best of our knowledge, this is the first work where it is used for the CZSL task.

\subsection{Masking}
Masking is commonly referring to a technique inspired by the Cloze task~\cite{Taylor1953Cloze} in the NLP community.
As in BERT~\cite{Devlin2019BERTPO}, it is commonly used as a proxy task to pretrain a language model where the model predicts the masked parts from the remaining part of a training sentence.
It is extended to multimodal language-vision models like ViLBERT~\cite{Lu2019vilbert,Lu2020MultitaskVL}.
While sharing similar design ideas, the proposed BMP mechanism is different from the token masks in prior works in the following aspects.
(1)~Objective: BMP brings consistent training and inference process for seen and unseen composite concepts, whereas the token mask improves modeling of the tokens' context;
(2)~Form: The edge blocking in BMP is applied methodically on the edges of the selected composite concepts and permanently changes the flow of information, whereas the token mask is usually randomly applied on tokens; and
(3)~Task: BMP is not used in pretraining tasks, while the token mask is mainly used in pretraining tasks.

%% file: depd/fig-framework-overview.tex
\begin{figure*}[!t]
    \centering
    \includegraphics[width=1.0\textwidth]{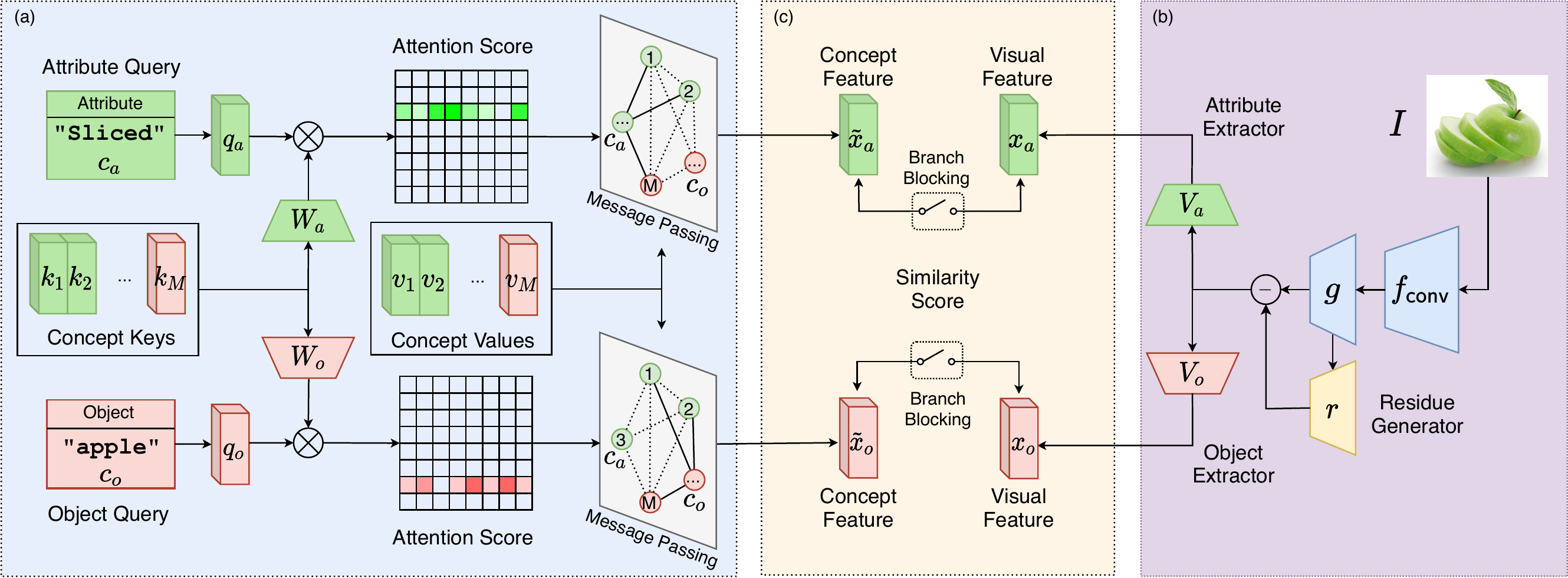}
    \vspace{-2em}
    \caption{\label{fig:framework-overview} 
        Overview of the proposed framework.
        Each colored block indicates a module.
        (a)~The \textit{concept module} generates features for the input attribute and object concept.
        A key-query based attention mechanism is used to capture the correlation between primitive concepts.
        (b)~The \textit{visual module} extracts visual feature from the input image.
        The primitive visual features for attribute and object concepts are extracted by two multi-layer perceptron $V_a$ and $V_o$.
        (c)~The \textit{matching module} calculates the similarity between attribute/object features from concept module and visual module separately.
        The figure is best viewed in color.
        }
    \vspace{-1ex}
\end{figure*}

%% file: sec_methodology.tex
\section{Methodology}
\label{sec:methodology}

\subsection{Problem Formulation}

In this work, we consider a world with $M$ primitive concepts {\small $\Set{M} = \{ c_1, c_2, \ldots, c_M \}$} and a set of composite concepts {\small $\Set{C} = \{ \langle c_a, c_o \rangle \} \; \forall \; a,o = 1,2,\ldots,M $}, where {\small $\langle c_a, c_o \rangle$} is a composite concept expressed as an ordered concept pair.\footnote{Having composite concepts with cardinality larger than two is possible (\eg~a \mbox{$\langle \mathsf{white}, \mathsf{wet}, \mathsf{dog} \rangle$}). We only consider attribute-object pairs in this work.}
For example, \CplxConcept{red}{tomato} is a composite concept built from primitive concept $\mathsf{red}$ and $\mathsf{tomato}$. 
Here, $ c_a $ indicates an attribute concept and $ c_o $ is an object concept. 
We use $c_{ao}$ as the shorthand for {$\langle c_a, c_o \rangle$}. 
We have a set {\small $\mathcal{A}$} for all attribute concepts and a set {\small $\mathcal{O}$} for all object concepts. 
While all the primitive concepts are seen by the model, some composite concepts are not seen until inference time.
There is a ``known'' composite concept set {\small $\Set{K} \subset \Set{C}$} and an ``unknown'' concept set {\small $\Set{U} = \Set{C} \setminus \Set{K}$}. 
We denote by {\small $\Set{I}_{ao} $} the set of images that corresponds to concept {\small $ c_{ao} $}. 
We also denote {\small $\Set{I}_\Set{K}$} and {\small $\Set{I}_\Set{U}$} as the sets of images corresponding to concepts in {\small $\Set{K}$} and {\small $\Set{U}$}, respectively. 
{\small $\Mat{I}_* \in \Set{I}_*$} is used to represent an image in any of the sets described above.
We use $\Vec{x}_*$ to denote visual feature from {\small $\Mat{I}_*$} and $\tVec{x}_*$ for concept features generated from {\small $c_*$}.
$x_{*,i}$ denotes the $i^{\text{th}}$ dimension of $\Vec{x}_*$.

The problem we solve is recognizing the composite concepts in {\small $\mathcal{I}_{\Set{U}}$} using the knowledge we have about {\small $\Set{I}_\Set{K}$}. 
More precisely, given an image {\small $\Mat{I}_{pq} \in \Set{I}_\Set{U}$}, the task is to correctly recognize the corresponding concept $c_{pq}$. 
Since {\small $\Mat{I}_{pq}$} and $c_{pq}$ are not seen by the framework until inference, we need to transfer the knowledge about primitive concepts in {\small $\Set{K}$} and {\small $\Set{I}_\Set{K}$} to recognize the composite concepts in {\small $\Set{I}_{\Set{U}}$}.

\subsection{Proposed Blocked Message Passing Network (BMP-Net)}

An overview of the BMP-Net is shown in \figref{fig:framework-overview}. 
There are two modality-based modules in our framework.
The first module is the {\bf concept module}, which takes a concept $c_i$ as input and produces $\tVec{x}_{i}$ as its concept feature.
The second module is the {\bf visual module}, which takes an image {\small $\Mat{I}$} and generates the corresponding primitive visual features of attribute $c_i$ and object $c_j$ in {\small $\Mat{I}$} as $\Vec{x}_{i}$ and $\Vec{x}_{j}$.
Two types of matching branches are used to match the representations produced by both modules.
Our framework recognizes the composite concept {\small $c_{pq}$} in image {\small $\Mat{I}_{ij}$} using the following steps:
\begin{enumerate}
    \item For each {\small $c_{pq} \in \Set{C}$}, use the concept module to generate primitive concept features $\tVec{x}_{p}$ and $\tVec{x}_{q}$. 
    \item Given an image {\small $\Mat{I}_{ij}$}, first use the visual module to extract its composite visual feature $\Vec{x}_{ij}$. 
    Then, two primitive visual features $ \Vec{x}_{i}$ and $\Vec{x}_{j} $ are extracted from $ \Vec{x}_{ij} $.
    \item A similarity score between $ c_{pq} $ and $ \Mat{I}_{ij} $ is calculated using the features generated in step 1 and 2.
    The composite concept that yields the highest similarity score is used as the recognition result.
\end{enumerate}

We will first discuss the message-passing mechanism in our concept module in Section~\ref{sssec:naive-attention} and Section~\ref{sssec:masked-attention}.
Following that, the visual module will be introduced in Section~\ref{sssec:vis-module}, and Section~\ref{sssec:branch-blocking} delineates the branch blocking.
Finally, we discuss the inference process in~Section~\ref{sssec:vis-infer}.

\subsubsection{Naive Message Passing}
\label{sssec:naive-attention}

As elaborated in Section~\ref{sec:introduction},
different primitive concepts have semantics that are related in different ways.
To model this correlation, we propose to condition the feature of a primitive concept on other primitive concepts. 
To fully exploit the semantics of primitive concepts and learn the potential co-relation between them, we use a key-query based attention mechanism to generate the primitive features.
Specifically, each primitive concept $c_p$ is described by a set of learnable parameters {\small $\langle \Vec{k}_p, \Vec{q}_p, \Vec{v}_p \rangle$}, which represents key, query, and value, respectively.
The key and the query contain the semantic information of a concept and are used to determine how two primitive concepts are related.
The relation is then used to guide the flow of messages (values) between concepts.

Concretely, given a primitive concept $c_p$, we first calculate the attention score between $c_p$ and all primitive concepts (including itself) using their keys and queries.
Assuming $c_p$ is an attribute concept, the attention score is calculated as
\begin{equation}
    \begin{split}\label{eq:naive-attention}
        \Vec{\alpha}_{p,a} &= \tanh(\Mat{W}_a \ \Mat{K}_a) \tanh(\Vec{q}_p) \\
        \Vec{\alpha}_{p,o} &= \tanh(\Mat{W}_a \ \Mat{K}_o) \tanh(\Vec{q}_p) \\
        \Vec{\beta}_p  &= \big[ \text{softmax}(\Vec{\alpha}_{p,a}) \ ; \ \text{softmax}(\Vec{\alpha}_{p,o}) \big]
    \end{split}
\end{equation}
where $\Vec{\alpha}_{p,a}$ ($\Vec{\alpha}_{p,o}$) is $c_p$'s un-normalized attention over all attribute (object) concepts, $\Vec{\beta}_p$ is the normalized attention score, $\Mat{K}_a$ and $\Mat{K}_o$ are the matrices containing stacked keys for all attribute and object concepts, and $\Mat{W}_a$ is the key transformation matrix for attributes.
We replace $\Mat{W}_a$ with $\Mat{W}_o$ if an object concept $c_q$ is given as input. 
The normalized attention vector $\text{softmax}(\Vec{\alpha})$ is calculated as
\begin{equation}
    \text{softmax}(\Vec{\alpha}) = \Bigg[ \frac{e^{\alpha_{1}}}{\sum_{j}^L e^{\alpha_{j}} }, \frac{e^{\alpha_{2}}}{\sum_{j}^L e^{\alpha_{j}} }, \ldots, \frac{e^{\alpha_{L}}}{\sum_{j}^L e^{\alpha_{j}} } \Bigg]^{\textsf{T}}
\end{equation}
where $L$ is the number of elements in $\Vec{\alpha}$.
Note that the normalization of attention scores in \eqref{eq:naive-attention} is performed on attribute and object concepts separately.
This is to prevent the model from focusing on a single type of primitive concept. 

The feature of primitive concept $c_p$ is generated by gathering the transformed values passed from other primitive concepts, \ie
\begin{equation}\label{eq:naive-attention-value}
    \tVec{x}_p = \text{LeakyReLU}\Bigg( \sum_{c_k \in \Set{M}} \beta_{p,k} \times (\Mat{U}_k \Vec{v}_k+\Vec{b}_k) \Bigg),
\end{equation}
where $\Vec{v}_k$, $\Mat{U}_k$ and $\Vec{b}_k$ are the value and transformation parameters of $c_k$.

\subsubsection{Edge Blocking}
\label{sssec:masked-attention}

The naive message passing (MP) causes a discrepancy between seen and unseen attribute-object pairs during inference.
When inferring a seen pair, the model can directly utilize the relation between the corresponding primitive concepts, which is well-established during training.
However, the process is not as direct when inferring an unseen pair because the relation between the corresponding primitive concepts is unknown.
Such discrepancy could severely harm the correctness of concept feature for unseen pairs, 
leading to a biased and sub-optimal model.
Moreover, the naive MP distributes an attribute (object)'s attention over all objects (attributes).
This could introduce noise in the concept feature as some attribute-object pairs never appear during training.

To overcome these problems, we propose a blocking mechanism on the edges between primitive concepts.
Given an input composite concept $c_{pq}$, the modified attention with edge blocking for attribute concept $c_p$ is calculated as
\begin{equation}
    \begin{split}\label{eq:masked-attention-attribute}
        \Vec{\alpha}^m_{p,a} &= \tanh(\Mat{W}_a \ \Mat{K}_a) \tanh(\Vec{q}_p) \\
        \alpha^m_{p,o,i} &= \begin{cases}
            -\infty,                                       & \!\!\!\! \text{if} \ c_{pi} \in \Set{U} \cup \{ c_{pq} \} \\
            \tanh(\Mat{W}_a \ \Mat{K}_o) \tanh(\Vec{q}_p), & \!\!\! \text{otherwise}\\
        \end{cases} \\
        \Vec{\beta}^m_p  &= \big[ \text{softmax}(\Vec{\alpha}^m_{p,a}) \ ; \ \text{softmax}(\Vec{\alpha}^m_{p,o}) \big] \\
    \end{split}
\end{equation}
whereas the modified attention for object concept $c_q$ is
\begin{equation}
    \begin{split}\label{eq:masked-attention-object}
        \Vec{\alpha}^m_{q,o} &= \tanh(\Mat{W}_o \ \Mat{K}_o) \tanh(\Vec{q}_q) \\
        \alpha^m_{q,a,i} &= \begin{cases}
            -\infty,                                       & \!\!\!\! \text{if} \ c_{iq} \in \Set{U} \cup \{ c_{pq} \} \\
            \tanh(\Mat{W}_o \ \Mat{K}_a) \tanh(\Vec{q}_q), & \!\!\! \text{otherwise}\\
        \end{cases} \\
        \Vec{\beta}^m_q  &= \big[ \text{softmax}(\Vec{\alpha}^m_{q,a}) \ ; \ \text{softmax}(\Vec{\alpha}^m_{q,o}) \big]
    \end{split}
\end{equation}

\input{depd/fig-mask-illustration.tex}

Intuitively, we block the edge before the normalization step if the primitive concepts involved are not seen pairs.
Moreover, the score related to the edge of training sample $c_{pq}$, \ie~$\alpha_{p,o,q}$ and $\alpha_{q,a,p}$, are also discarded.
By doing so, we gear the model to use the seen neighboring primitive concepts to infer the input composite concept $c_{pq}$.
On the one hand, this removes the noise introduced by the attention scores between unseen pairs.
On the other hand, this equalizes the inference process for seen and unseen pairs, where the advantage of seen pairs gained from direct information flow no longer exists.

An illustration of the naive and blocked MP scheme is shown in \figref{fig:mask-illustration} using a mini-world of 5 primitive concepts, where the composite concepts \CplxConcept{1}{5} and \CplxConcept{3}{4} are unseen during training.
As shown in \figref{fig:mask-illustration}-a, it is noticeable that information will not flow between unseen pairs when edge blocking is applied, which reduces noise during training.
The discrepancy between seen and unseen pairs during training is shown in \figref{fig:mask-illustration}{-b/c} with the naive MP scheme.
Since $c_{15}$ is not seen by the model until inference time, the relation between $c_1$ and $c_4$ could be excessively strengthened during training, which suppresses the attention between $c_1$ and $c_5$ and leads to a sub-optimal result on the unseen pair $c_{15}$ during inference.
The problem is overcome by blocking the edge between input primitive concept pairs.
As depicted in \figref{fig:mask-illustration}{-b/c}, the information does not flow directly between the input primitive concepts (\ie~$c_{1}$ or $c_{5}$) with the blocked MP scheme, but between them and their neighbors.
In this way, the model will infer $c_{14}$ and $c_{15}$ in a consistent way and not accumulate bias on the seen pair $c_{14}$ during training.

Similar to \eqref{eq:naive-attention-value}, the primitive concept feature for $c_p$ is calculated as
\begin{equation}
    \begin{split}\label{eq:masked-attention-value}
        \tVec{x}^m_p &= \text{LeakyReLU}\Bigg( \sum_{c_k \in \Set{M}} \beta^m_{p,k} \times (\Mat{U}_k \Vec{v}_k+\Vec{b}_k) \Bigg) \\
    \end{split}
\end{equation}

While edge blocking helps solve the problems in naive MP, it also ignores a large number of connections between primitive concepts (\ie~the unseen pairs).
To make the most of both naive and blocked schemes, we combine them together during training.
This is detailed in Section~\ref{sssec:combine-naive-masked}.

\subsubsection{Visual Module}
\label{sssec:vis-module}
Given an input image $\Mat{I}_{ij}$, the visual module first transforms its visual feature into a composite feature $\Vec{x}_{ij}$.
The problem with features output by the ConvNet backbone $f_\textsf{conv}$ is that images inevitably contain non-discriminative information, which is relevant to neither the attribute nor the object.
We call such information the \textit{residue}.
Residues do not improve model training but could lead to over-fitting.
Inspired by the image denoising literature~\cite{Zhang2017BeyondGD,Fu2017RemovingRainSI}, we introduce a network that learns to remove residues from the input visual features.
The residue component can be regarded as noise in the image feature and subtracted from the image feature to improve the training process.
This step can be formulated as follows:
\begin{equation}\label{eq:vis_feature}
    \begin{split}
        \Vec{x}'_{ij} &= g \big( f_\textsf{conv}(\Mat{I}_{ij}) \big) \\
        \Vec{x}_{ij} &= \Vec{x}'_{ij} - r( \Vec{x}'_{ij} )
    \end{split},
\end{equation}
where $f_{\textsf{conv}}$ is the ConvNet backbone, $g()$ is a learnable transformation network, and $r()$ is the residue generator.
The primitive visual features for attribute $c_i$ and object $c_j$ are extracted %
using two MLPs as {\small $\Vec{x}_i = V_a(\Vec{x}_{ij})$} and {\small $\Vec{x}_j=V_o(\Vec{x}_{ij})$}.

In our implementation, we assume the residue can be sampled from a multi-variate Gaussian distribution with learnable parameters.
In the training phase, the residue is sampled from this distribution. 
The parameters of the residue generator are optimized by backpropagation through the reparameterization trick~\cite{Kingma2013AutoVB}.
In the inference phase, the mean of this distribution is directly taken as the residue.

\subsubsection{Branch Blocking}\label{sssec:branch-blocking}
This section discusses the second usage of blocking, which reduces the entanglement between attributes and objects.
Since attributes and objects are highly intertwined in the real world, it is imminent that the attribute information will introduce bias to the object branch during training and vice versa.
As a result, both branches are prone to overfitting and fail when encountering a new attribute-object pair.
This phenomenon is similar to feature co-adaptation~\cite{Hinton2012ImprovingNN}, which causes networks to overfit.
A cure to curtail the interdependence between the attribute and the object is to reduce their co-occurrence during training.
Inspired by dropout~\cite{Hinton2012ImprovingNN} and path-drop~\cite{Larsson2017Fractalnet}, we disable the attribute (object) branch during some randomly determined iterations so that only the object (attribute) branch is trained.
More specifically, before each training iteration, we generate two random numbers {\small $\omega_a, \omega_o \sim U(0,1)$}.
For a threshold $\tau$, if $\omega_a < \tau$, the whole attribute branch will be blocked, and the model will be only trained to recognize objects.
The same procedure is also applied to the object branch using $\omega_o$.

The branch blocking and edge blocking together forms our BMP method.
The edge blocking maintains the consistency between seen and unseen pairs by modifying the connection between primitive concepts, while the branch blocking decouples attributes and objects in the training process.

\subsubsection{Inference Procedure}\label{sssec:vis-infer}

The similarity score between input image {\small $\Mat{I}_{ij}$} and candidate composite concept $c_{pq}$ is defined as
\begin{equation*}\label{eq:simscore-separate}
    \text{sim} \big( \Mat{I}_{ij}, c_{pq} \big) = -d \big( \Vec{x}_i, \tVec{x}^m_p \big) - d \big( \Vec{x}_j, \tVec{x}^m_q \big),
\end{equation*}
where {\small $d(\cdot, \cdot)$} is the Euclidean distance between two vectors.
Finally, the recognition result is generated as
\begin{equation}\label{eq:separate-match}
    c^* =\underset{c_{pq}}{\argmax} \ \text{sim} \big( \Mat{I}_{ij}, c_{pq} \big).
\end{equation}

\vspace{-3ex}
\subsection{Training}\label{ssec:training}

In this work, our goal is to train a network to correctly recognize the composite concepts in the given input images.
To align this goal with our inference procedure, we train the model to minimize the distance between features of the corresponding images and concepts. Towards this goal, in each training iteration, we sample a mini-batch from {\small $\Set{I}_\Set{K}$}. 
Each sample in the mini-batch consists of one reference image-concept pair {\small {$\langle \Mat{I}_{a o}, c_{a o} \rangle$}} and two negative image-concept pairs, {\small {$\langle \Mat{I}_{\bar{a} o}, c_{\bar{a} o} \rangle$}} and {\small {$\langle \Mat{I}_{a \bar{o}}, c_{a \bar{o}} \rangle$}}, where {\small $\bar{a} \neq a$} and {\small $\bar{o} \neq o$}.
Note that we intentionally sample negative concepts that share the same object or attribute with the reference concept.

\subsubsection{Loss Functions}\label{sssec:loss-functions}

Before detailing our training scheme, we first introduce a function $l_t$ that is related to our defined loss functions.
This function is defined as
\begin{equation*}
l_t(\Vec{x}_n, \Vec{x}_p, \Vec{x}_r, m)=\log\Big(1+e^{ m-\big( d(\Vec{x}_r, \Vec{x}_n) - d(\Vec{x}_r, \Vec{x}_p)\big) } \Big),
\end{equation*}
which encourages a reference vector \( \Vec{x}_r \) to be close to the ``positive'' vector \(\Vec{x}_p\) and far from the ``negative'' vector \(\Vec{x}_n\) under distance function $d(\cdot, \cdot)$.
This function has a hyperparameter $m$, which is the margin between the distances.

\vspace{0.5ex}
\noindent\textbf{The Hinge Loss:} 
In \eqref{eq:separate-match}, the Euclidean distance between features plays a vital role during inference stage. 
The matching image-concept feature pairs should appear close to each other in the feature space.
Therefore, the essential training objective is to push the features of different concepts away from each other.
Formally, the objective with margin $m_h$ is defined by the following loss functions
\begin{equation}\label{eq:hinge_loss}
    \begin{split}
        \mathcal{L}_\textsf{v} &= l_t(\tVec{x}^m_{\bar{a}}, \tVec{x}^m_{a}, \Vec{x}_a, m_h) + l_t(\tVec{x}^m_{\bar{o}}, \tVec{x}^m_{o}, \Vec{x}_o, m_h) \\
        \mathcal{L}_\textsf{c} &= l_t(\Vec{x}_{\bar{a}}, \Vec{x}_{a}, \tVec{x}^m_a, m_h) \; + l_t(\Vec{x}_{\bar{o}}, \Vec{x}_{o}, \tVec{x}^m_o, m_h)
    \end{split}
\end{equation}
The difference between the two hinge loss terms is that $\mathcal{L}_\textsf{v}$ focuses more on the visual features while $\mathcal{L}_\textsf{c}$ focuses more on the concept features.
In practice, we found that both terms are important for the model to learn well.

\vspace{0.5ex}
\noindent\textbf{The Classification Loss:} 
It has been shown in previous works~\cite{Nagarajan2018AttributesAO, Wei2019AdversarialFC} that auxiliary classification losses can help make attribute/object features more discriminative. 
We anticipate the same in our framework.
To this end, we feed the extracted attributes and objects feature to train two softmax classifiers, and minimize the following negative log-likelihood loss:
\begin{equation}\label{eq:loss-decons-cls}
    \begin{split}
    \mathcal{L}_\textsf{aux} =
    &\sum_{c_k \in \Set{A}} - \mathbb{I} \big( c_{a} = c_k \big) \times \log\big(p_{a}(c_k | \tVec{x}^m_{a})\big) \: + \\
    &\sum_{c_k \in \Set{O}} - \mathbb{I} \big( c_{o} = c_k \big) \times \log\big(p_{o}(c_k | \tVec{x}^m_{o})\big),
    \end{split}
\end{equation}
where {\small $\mathbb{I}(\cdot)$} is the indicator function.
$p_a$ ($p_o$) is the attribute (object) probabilities given by the auxiliary classifier.

\vspace{0.5ex}
\noindent\textbf{The Combined Loss:}
The final loss function is defined as
\begin{equation}
    \mathcal{L} = 
    \lambda_\textsf{v} \mathcal{L}_\textsf{v} + 
    \lambda_\textsf{c} \mathcal{L}_\textsf{c} + 
    \lambda_\textsf{a} \mathcal{L}_\textsf{aux},
\end{equation}
where $\lambda_*$ are hyperparameters that balances different terms.

\subsubsection{Combined Naive MP and Blocked MP}\label{sssec:combine-naive-masked}

As discussed in Section~\ref{sssec:masked-attention}, both MP schemes have advantages and drawbacks.
Specifically, there is a clear trade-off between better exploitation of relations and reducing bias on seen pairs.
To make the best of these two schemes, we propose to train the model using both mechanisms.
During training,
we generate $\tVec{x}_*$ and $\tVec{x}^m_*$ using both naive and blocked MP from the training samples.
The loss function being optimized is
\begin{equation}\label{eq:combine-naive-blocked}
    \mathcal{L}' = \mathcal{L} + \lambda_\textsf{r} \mathcal{L}_\textsf{r},
\end{equation}
where the reconstruction term {\small $\mathcal{L}_r = \Norm{\tVec{x}^m_a - \tVec{x}_a}^2_2 + \Norm{\tVec{x}^m_o - \tVec{x}_o}^2_2 $} improves the consistency between concept features generated from naive and blocked MP.
In this way, the model can leverage the information exploited by naive MP without changing the BMP-based inference procedure.
Note that the inclusion of naive MP does not change the inference process, where only blocked MP is used.

%% file: depd/fig-mask-illustration.tex
\begin{figure}[!t]
    \centering
    \includegraphics[width=0.92\columnwidth]{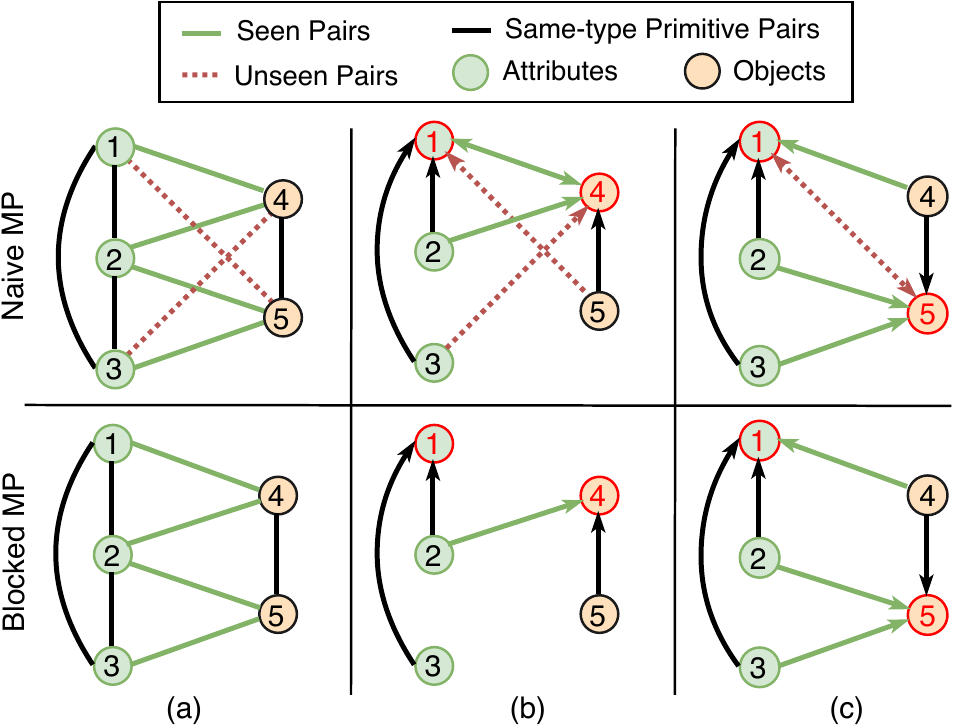}
    \vspace{-1em}
    \caption{\label{fig:mask-illustration}
      Illustration of naive and blocked MP scheme in a mini-world of 3 attribute and 2 object concepts,
      where the composite concepts \CplxConcept{1}{5} and \CplxConcept{3}{4} are unseen during training.
      The queried concepts are marked in {\color{red} red} and arrows indicate the directions of information flows.
      (a) shows the relationships used during attention calculation.
      (b) and (c) show the flow of information between primitive concepts when the input composite concept is seen and unseen, respectively.
    }
\end{figure}

%% file: sec_experiment.tex
\section{Experiment}\label{sec:experiment}

\input{depd/table-auc}

\subsection{Experimental Setup}

\noindent\textbf{Datasets.}
We conduct experiments on two widely used benchmark datasets, namely UT-Zappos~\cite{Yu2014FineV} and MIT-States~\cite{Isola2015DS}.
The detailed configuration of the datasets is provided in~\cite{Purushwalkam2019TaskDrivenMN}.

UT-Zappos~\cite{Yu2014FineV} has about 33,000 shoe images with a white background.
The attribute concepts are mainly the material (\eg~\PrimConcept{rubber}) while the object concepts are the types of shoes (\eg~sandals).
There are 12 object concepts and 16 attribute concepts. 
In total, there are 83 composite concepts in {\small $\Set{K}$} and 33 composite concepts in {\small $\Set{U}$}.
The training set has about 23,000 images.
The validation set has about 3,000 images from 15 seen and 15 unseen composite concepts.
The test set has about 3,000 images from 18 seen and 18 unseen composite concepts.

MIT-States~\cite{Isola2015DS} has about 53,000 images with 245 object concepts and 115 attribute concepts.
It is more challenging than UT-Zappos due to its larger scale and higher visual diversity and covers a great range of daily life composite concepts (\eg~\CplxConcept{blunt}{knife} and \CplxConcept{old}{house}).
There are 
1,262 composite concepts in {\small $\Set{K}$} and 700 composite concepts in {\small $\Set{U}$}.
The training set has about 30,000 images.
The validation set has about 10,000 images from 300 seen and 300 unseen composite concepts.
The test set has about 13,000 images from 400 seen and 400 unseen composite concepts.

\vspace{0.5ex}
\noindent\textbf{Metrics.}
The original evaluation protocol used by most CZSL works was proposed in~\cite{Nagarajan2018AttributesAO}.
Purushwalkam~\etal~\cite{Purushwalkam2019TaskDrivenMN} improved this protocol with a validation set and a procedure to better reflect the overall performance of the system.
In this work, we adopt the improved evaluation protocol~\cite{Purushwalkam2019TaskDrivenMN}.
Briefly, a scalar called calibration bias is added to the similarity scores between unseen composite concepts and the input image.
For a given calibration bias, the system's accuracy on seen concepts and unseen concepts is calculated.
As the calibration bias varies, the model goes from focusing purely on seen concepts to purely on unseen concepts, and an unseen-seen accuracy curve can be drawn (see \figref{fig:auc_figs} for example).
Area Under Curve (AUC) is adopted to show the overall performance of the system. 
For complementary comparison with the classic evaluation protocol~\cite{Nagarajan2018AttributesAO}, \ie~the best accuracy on seen/unseen pairs and the best harmonic mean of seen and unseen accuracy (we term it calibrated H-Mean, or cH-Mean, in this paper), is also reported.
All the accuracies are calculated for the top-1,2,3 predictions.
We refer the readers to~\cite{Purushwalkam2019TaskDrivenMN} for details.

\vspace{0.5ex}
\noindent\textbf{Implementation.}
For a fair comparison, we follow~\cite{Nagarajan2018AttributesAO, Purushwalkam2019TaskDrivenMN} and use ResNet-18~\cite{He2016DeepRL} pretrained on ImageNet as the backbone network $f_{\mathsf{conv}}$ in all experiments. 
The backbone is not finetuned during training.
We use the 512-dimension output of the last fully connected layer of $f_{\mathsf{conv}}$ as the input visual feature.
All the leaky ReLU activation functions use a negative slope of $0.1$.
Both concept and visual features are in $\mathbb{R}^{512}$.
We use a batch size of 512 for both datasets.
The training takes about 5 hours for MIT-States and about 40 minutes for UT-Zappos on two NVIDIA V100 GPUs.
All the hyper-parameters are empirically determined by optimizing the model's performance on the validation set.
We set $m=0.5$ as the margin in \eqref{eq:hinge_loss}, and $\tau=0.05$ for branch blocking. 
The weights $\{ \lambda_\textsf{v}, \lambda_\textsf{c}, \lambda_\textsf{a}, \lambda_\textsf{r} \}$ are $\{ 10, 0.5, 1, 10 \}$ for UT-Zappos and $\{ 20, 5, 10, 5 \}$ for MIT-States.
Our code and trained model is available at \url{https://github.com/daoyuan98/Relation-CZSL}.

\vspace{-3ex}
\subsection{Recognizing Images with Composite Concepts}

\noindent\textbf{Baselines.}
We compare against the following works:

\noindent(1)~AttrAsOp~\cite{Nagarajan2018AttributesAO}:
An attribute $ c_i $ is modeled as a transformation matrix $\Mat{A}_i$ and an object $ c_j $ is modeled as a vector $ \Vec{o}_j $. 
The vectors of objects are transformed by the attribute in the composition process {\small $\tVec{x}_{ij} = \Mat{A}_i\Vec{o}_j$}. 
The output vector $ \tVec{x}_{ij} $ is used to match with the visual feature of the input image.

\noindent(2)~RedWine~\cite{Misra2017FromRW}: 
Each primitive concept is modeled as the weight of its corresponding linear classifier. 
A transform function is learned to merge these weights into a linear classifier used to classify composite concepts. 

\noindent(3)~LabelEmbed+~\cite{Nagarajan2018AttributesAO}:
The attribute and object concepts are represented by GloVe~\cite{Pennington2014Glove} word vectors and then concatenated.
The concatenated concept vectors and the visual feature vectors are embedded into a common feature space.

\noindent(4)~FeatureGen~\cite{Xian2017FeatureGN}: 
This work proposes to train a Wasserstein GAN to generate features for the unseen concepts.
The generated features are used to train softmax classifiers that work well on unseen concepts.

\noindent(5)~TMN~\cite{Purushwalkam2019TaskDrivenMN}: 
TMN proposes a task-driven modular network architecture where the compositional reasoning task is divided into sub-tasks solvable by many small networks operating in a semantic concept space.
Reweighted responses of those small networks are used to generate the recognition results.

\noindent(6)~SymNet~\cite{Li2020SymmetryGA}: 
This work exploits the symmetry principle in the attribute-object composition process and builds a transformative framework inspired by group theory.
A relative moving distance is also introduced to better recognize attributes. 

\noindent(7)~AFGCL and AFGCL-960~\cite{Wei2019AdversarialFC}: 
A GAN is trained with a delicate quintuplet loss which enforces fine-grained structure within the feature space.
Relations between attributes and objects are partially modeled by the loss.
Same as other baseline methods, the plain AFGCL is trained using 512-dimension ResNet-18 features.
The AFGCL-960 is the improved version that uses 960-dimension multi-scale ResNet-18 features.

The results for baseline (1)-(5) are reported by~\cite{Purushwalkam2019TaskDrivenMN}, whereas the results for baseline (6) and (7) are provided by the authors.

\input{depd/table-ablation-study}

\input{depd/table-hmean}

\vspace{0.5ex}
\noindent\textbf{Results.}
We follow the settings in~\cite{Purushwalkam2019TaskDrivenMN} to evaluate our model and compare with the state-of-the-art methods.
\tabref{table:auc} shows our model's AUC compared with baselines on both validation set and test set. 
In both datasets, our model outperformed baseline methods by a large margin, showing the benefit of relation-aware learning of the concept module.
Our model achieves further improvement when using the 960-dimension multi-scale feature, outperforming its counterpart AFGCL-960.
\tabref{table:acc} shows the accuracy on seen and unseen composite concepts, as well as the best cH-Mean of the two when the bias varies.
On UT-Zappos, our model is able to outperform baselines on seen composite concepts while achieving competitive accuracy on unseen ones.
On MIT-States, our method achieves better overall performance than all baselines methods.
As shown by the unseen-seen accuracy curve in \figref{fig:auc}, our method keeps a better balance between seen and unseen pairs, which leads to better performance.
For the sake of completeness, we also provide the top-1,2,3 accuracy of our model in \figref{fig:topk}.

\vspace{0.5ex}
\noindent\textbf{Ablation Study.}
We examine the effects of various components on our framework.
the hinge loss terms {\small $\lambda_{\textsf{v}/\textsf{c}}$}, the auxiliary classifiers, the reconstruction term $\mathcal{L}_r$ combining the naive/blocked MP in~\eqref{eq:combine-naive-blocked}, the blocking mechanism, and the residue generator.
The results are shown in \tabref{table:ablation-study}.
The first remark is that the removal of any component from the proposed framework generally results in a worse performance (measured by test AUC and cH-Mean).
Among all the loss terms, $\lambda_\textsf{v}$ plays a particularly important role as it constrains the structures in the rich visual features.
Overall speaking, the result upholds the efficacy of our framework design.
On the other hand, it is noticeable that on MIT-States, the model's unseen accuracy drops with the use of blocking,
which is seemingly in contrast to its design intentions.
However, it should be noted that the model's performance should be comprehensively interpreted using various calibration biases.
As shown in \figref{fig:auc_ablation}, if we compare the unseen-seen curve of our full model and the variant without blocking, it is obvious that the former outperforms the latter for most of the calibration biases in terms of unseen accuracy.
This also results in the better test AUC and cH-Mean of the full model.
Therefore, we conclude that the masked attention has arguably improved the model's performance on unseen pairs, and therefore the overall performance.

As introduced in~\eqref{eq:combine-naive-blocked}, the blocked MP is used together with naive MP controled by hyperparameter $\lambda_\textsf{r}$.
Intuitively, $\lambda_\textsf{r}=0$ results in a pure blocked MP model, and a larger $\lambda_\textsf{r}$ indicates a stronger influence of naive MP\@.
To examine the importance of naive MP, we train our model with different $\lambda_\textsf{r}$ values.
As shown in \tabref{table:ablation-lbdr}, using only blocked MP results in an inferior performance.
This is because the edges between unseen pairs are discarded during training, which results in incomplete exploitation of relations between primitive concepts.
This issue is alleviated by the inclusion of naive MP due to its complete (but bias-inducing) modeling of primitive concepts during training.

\input{depd/fig-auc}
\input{depd/fig-topk}

\input{depd/table-ablation-lbdr}

\input{depd/fig-feat-vis}

\vspace{0.5ex}
\noindent\textbf{What has the model learned?}
Since the model recognizes attribute and object concepts from visual inputs by matching the concept and visual features, the question can be answered by finding out the visual correspondences of the learned primitive concept features.
Inspired by prior works on deep network feature visualization~\cite{Dosovitskiy2016InvertingVR}, we start from a randomly initialized image $\Mat{I}$ and a target primitive concept feature $\tVec{x}_{a/o}^m$, and solve the following optimization problem
\[ \Mat{I}^* ={\arg\min}_\Mat{I} || V_{a/o}(\Vec{x}_\Mat{I}) - \tVec{x}_{a/o}^m ||_2^2 + \lambda (|| \Delta_x \Mat{I} ||_2^2 + || \Delta_y \Mat{I} ||_2^2), \]
where {\small $\Vec{x}_\Mat{I}$} is the primitive visual feature of image {\small $\Mat{I}$} extracted following~\eqref{eq:vis_feature} and {\small $\Delta_{x/y} \Mat{I}$} is the first order difference of the image along its axis.
The optimization is solved by freezing the model parameters and iteratively updating {small $\Mat{I}$} using gradient descent.
We can also visualize the residual component by replacing {\small $\Mat{x}_{a/o}^m$} with the residual component, and {\small $V_{a/o}(\Vec{x}_\Mat{I})$} by {\small $f_{\textsf{conv}}(\Mat{I})$}.

As shown in \figref{fig:feat-vis}, the resultant images {\small $\Mat{I}^*$} provide cues to the different ways the attributes and objects are perceived by the model.
The attributes can be identified by its geometric characteristics (\eg~\PrimConcept{winding}, which contains many curved lines, and \PrimConcept{diced}, which contains groups of small spots), colors (\eg~\PrimConcept{verdant}), or both (\eg~\PrimConcept{barren}, which has a brownish color and dark spots indicating probably withered plants).
Such properties are not related to specific objects, as we can hardly see any interpretable shapes for object concepts.
Objects, on the other hand, are mainly defined by the outline of distinctive components (\eg~head/legs for the \PrimConcept{iguana}).
Such an outline is generally independent of attributes.
For example, we can see \PrimConcept{tower}s in different scales and straight or curved \PrimConcept{highway}s.
Finally, the visualization of residual components contains mainly irregular shapes composed of random lines and blobs.
This aligns with its design goal, and we interpret the result as a collection of low-level features that is less helpful for identifying attributes and objects.
All these observations indicate that our model can effectively learn decoupled and discriminative features separately for attributes and objects.

\input{depd/fig-failure-analysis}

\input{depd/fig-retrieval}

\vspace{0.5ex}
\noindent\textbf{Failure case analysis.}
\revtextB{
We show some failure cases in \figref{fig:failure-analysis} for a better understanding of the dataset and our model.
From these examples, we identify two causes of failures.
The first cause is that multiple concepts can be used to describe a single image.
For example, \PrimConcept{fabric}-\PrimConcept{clothes}, \PrimConcept{mossy}-\PrimConcept{verdant}, and \PrimConcept{copper}-\PrimConcept{brass} are interchangeable in the shown cases.
This calls for better annotation of the dataset and more flexible evaluation protocols.
The second cause is visual ambiguity.
For example, in \figref{fig:failure-analysis}(d), the cupboard is misclassified as bookshelves, which results in the wrong prediction of objects.
This could be improved by better backbone models.
As a side note, our model predicts both seen and unseen pairs in a balanced manner due to the blocked MP, which reduces biases on seen ones.
}

\vspace{0.5ex}
\noindent\textbf{Qualitative evaluation via image retrieval.}
We examine how our model exploits relations between primitive concepts and generalizes to unseen composite concepts on the image retrieval task.
Two scenarios are considered.
First is the ``unseen'' concept, where the images for these concepts are provided by the dataset but not used by the model during training.
The second is the ``out-of-the-world'' concept.
Concepts of this type are in set $\Set{C}$, but are not supported by any image in the dataset (\eg~\CplxConcept{cooked}{tree} does not exist in the dataset).

The retrieval results with MIT-States are shown in \figref{fig:retrieval}, where we compare our model with TMN~\cite{Purushwalkam2019TaskDrivenMN}.
Our model is able to correctly retrieve images for unseen concepts like \CplxConcept{windblown}{tree} and \CplxConcept{rusty}{truck}.
Some of the reported errors are due to the object ambiguity (\eg~(b1) can also be labeled as \CplxConcept{ruffled}{chair}) or attribute ambiguity (\eg~in (c5) the visual difference between \PrimConcept{rusty} and the ground-truth \PrimConcept{burnt} is small).
For out-of-the-world concepts, with proper exploitation of relations between primitive concepts, our model is able to retrieve visually or semantically similar results.
For example, the results for \CplxConcept{weathered}{car} on (d) are mostly broken cars/trucks.
More interestingly, when given composite concepts with antonymous attributes, our model is returning results with opposite visual impressions or semantics.
For example, results for \CplxConcept{raw}{tree} and \CplxConcept{cooked}{tree} contain attribute pairs like \PrimConcept{verdant}-\PrimConcept{barren}.
The results for \PrimConcept{raw} are generally visually greenish, while the results for \PrimConcept{cooked} are mostly yellowish and show a hint of dryness.
In comparison, the results of TMN are more biased towards seen concepts and less accurate or related to the query.
This provides evidence that our model can better utilize the relations between primitive concepts to make reasonable associations.

%% file: depd/table-auc.tex
\begin{table*}[!t]
    \centering
    \caption{\label{table:auc}
        Top-k AUC ($\times 100$) on evaluated datasets.
        Best results are in bold.}
    \vspace{-1em}
    \adjustbox{width=\textwidth}{
    \begin{tabular}{l C{6ex} C{6ex} C{6ex} c C{6ex} C{6ex} C{6ex} c C{6ex} C{6ex} C{6ex} c C{6ex} C{6ex} C{6ex}} 
    \toprule
        & \multicolumn{7}{c}{MIT-States}                                                    & \multicolumn{1}{l}{} & \multicolumn{7}{c}{UT-Zappos}                                                     \\ \cmidrule{2-8} \cmidrule{10-16} 
        & \multicolumn{3}{c}{Val AUC} & \multicolumn{1}{l}{} & \multicolumn{3}{c}{Test AUC} & \multicolumn{1}{l}{} & \multicolumn{3}{c}{Val AUC} & \multicolumn{1}{l}{} & \multicolumn{3}{c}{Test AUC} \\ \cmidrule{2-4} \cmidrule{6-8} \cmidrule{10-12} \cmidrule{14-16} 
    Model k $\rightarrow$    & 1          & 2           & 3           & & 1          & 2           & 3           & & 1           & 2           & 3           & & 1           & 2           & 3           \\ \midrule
    AttrAsOp~\cite{Nagarajan2018AttributesAO} & 2.5    & 6.2         & 10.1        & & 1.6        & 4.7         & 7.6         & & 21.5        & 44.2        & 61.6        & & 25.9        & 51.3        & 67.6        \\
    RedWine~\cite{Misra2017FromRW}        & 2.9        & 7.3         & 11.8        & & 2.4        & 5.7         & 9.3         & & 30.4        & 52.2        & 63.5        & & 27.1        & 52.6        & 68.8        \\
    LabelEmbed+~\cite{Nagarajan2018AttributesAO} & 3.0 & 7.6         & 12.2        & & 2.0        & 5.6         & 9.4         & & 26.4        & 49.0        & 66.1        & & 25.7        & 52.1        & 67.8        \\
    FeatureGen~\cite{Xian2017FeatureGN}   & 3.1        & 6.9         & 10.5        & & 2.3        & 5.7         & 8.8         & & 20.1        & 45.1        & 61.1        & & 25.0        & 48.2        & 63.2        \\
    TMN~\cite{Purushwalkam2019TaskDrivenMN} & 3.5      & 8.1         & 12.4        & & 2.9        & 7.1         & 11.5        & & 36.8        & 57.1        & 69.2        & & 29.3        & 55.3        & 69.8        \\
    AFGCL~\cite{Wei2019AdversarialFC}     & 4.2        & 10.9        & 17.7        & & 3.9        & 9.9         & \best{15.9} & & 38.1        & 69.8        & 84.9        & & 37.8        & 67.2        & 81.4        \\
    SymNet~\cite{Li2020SymmetryGA}        & 4.3        & 9.8         & 14.8        & & 3.0        & 7.6         & 12.3        & & -           & -           & -           & & -           & -           & -           \\
    BMP-Net                               & \MErr{\best{6.0}}{0.1} & \MErr{\best{13.3}}{0.2} & \MErr{\best{19.6}}{0.1} & & \MErr{\best{4.3}}{0.1} & \MErr{\best{10.0}}{0.1} & \MErr{15.1}{0.2}        & & \MErr{\best{51.1}}{0.5} & \MErr{\best{74.0}}{0.3} & \MErr{\best{86.2}}{0.2} & & \MErr{\best{44.7}}{0.6} & \MErr{\best{72.5}}{0.5} & \MErr{\best{85.9}}{0.1} \\ \midrule
    AFGCL-960~\cite{Wei2019AdversarialFC} & 5.3        & 13.1        & 20.3        & & 4.0        & 10.4        & 16.4        & & 40.6        & 71.6        & 85.7        & & 39.5        & 69.7        & 82.5        \\
    BMP-Net-960                           & \MErr{\best{8.2}}{0.1} & \MErr{\best{17.2}}{0.2} & \MErr{\best{24.1}}{0.2} & & \MErr{\best{6.0}}{0.1} & \MErr{\best{13.3}}{0.1} & \MErr{\best{19.4}}{0.1} & & \MErr{\best{57.4}}{0.6} & \MErr{\best{78.6}}{0.3} & \MErr{\best{89.4}}{0.4} & & \MErr{\best{49.7}}{0.5} & \MErr{\best{76.8}}{0.4} & \MErr{\best{89.1}}{0.3} \\ \bottomrule
    \end{tabular}
    }
\end{table*}

%% file: depd/table-ablation-study.tex
\begin{table*}[!b]
    \centering
    \caption{\label{table:ablation-study}
        Ablation study with various model configurations.}
    \vspace{-1em}
    \adjustbox{width=\textwidth}{
    \begin{tabular}{l c ccccc c ccccc}
        \toprule
                                      & & \multicolumn{5}{c}{MIT-States}           &   & \multicolumn{5}{c}{UT-Zappos}                 \\
        \cmidrule{3-7} \cmidrule{9-13}
        Model                          & & Val AUC & Test AUC & Seen  & Unseen & cH-Mean & & Val AUC & Test AUC & Seen  & Unseen & cH-Mean \\
        \midrule
        w/o $\mathcal{L}_\textsf{v}$   & & \MErr{ 1.63}{0.08}    & \MErr{0.89}{0.09}     & \MErr{15.29}{0.78} & \MErr{ 8.62}{0.67}  & \MErr{ 6.83}{0.40} &  & \MErr{26.35}{5.55}   & \MErr{22.94}{3.43}    & \MErr{72.90}{2.41} & \MErr{37.78}{5.10}  & \MErr{37.27}{3.54}   \\
        w/o $\mathcal{L}_\textsf{c}$   & & \MErr{6.05}{0.08}    & \MErr{4.23}{0.07}     & \MErr{31.66}{0.46} & \MErr{18.73}{0.10}  & \MErr{15.63}{0.19} &  & \MErr{49.82}{1.15}   & \MErr{42.87}{0.88}    & \MErr{83.77}{0.59} & \MErr{59.30}{0.89}  & \MErr{54.44}{0.87}   \\
        w/o $\mathcal{L}_\textsf{aux}$ & & \MErr{5.06}{0.08}    & \MErr{3.46}{0.12}     & \MErr{29.12}{0.84} & \MErr{17.33}{0.21}  & \MErr{13.95}{0.20} &  & \MErr{50.98}{0.81}   & \MErr{44.03}{0.74}    & \MErr{83.58}{0.22} & \MErr{59.73}{0.83}  & \MErr{56.22}{0.59}   \\
        w/o $\mathcal{L}_r$            & & \MErr{5.77}{0.12}    & \MErr{3.90}{0.11}     & \MErr{30.10}{0.77} & \MErr{18.25}{0.15}  & \MErr{14.93}{0.22} &  & \MErr{50.10}{0.71}   & \MErr{43.52}{0.92}    & \MErr{84.07}{0.34} & \MErr{59.31}{1.09}  & \MErr{55.18}{1.03}    \\
        w/o Blocking                   & & \MErr{5.86}{0.14}    & \MErr{4.21}{0.07}     & \MErr{30.35}{0.23} & \MErr{19.41}{0.15}  & \MErr{15.33}{0.13} &  & \MErr{49.24}{1.03}   & \MErr{41.03}{0.96}    & \MErr{82.70}{0.47}  & \MErr{56.48}{0.98}  & \MErr{53.83}{0.82}   \\
        w/o Residue                    & & \MErr{5.82}{0.08}    & \MErr{4.01}{0.06}     & \MErr{31.44}{0.33} & \MErr{18.37}{0.17}  & \MErr{15.35}{0.24} &  & \MErr{52.10}{0.66}   & \MErr{44.27}{0.51}    & \MErr{83.75}{0.17} & \MErr{59.74}{0.65}  & \MErr{56.32}{0.51}   \\
        Full                           & & \MErr{6.01}{0.07}    & \MErr{4.31}{0.09}     & \MErr{32.53}{0.42} & \MErr{18.86}{0.21}  & \MErr{16.01}{0.18} &  & \MErr{51.12}{0.47}   & \MErr{44.69}{0.60}    & \MErr{83.91}{0.20} & \MErr{60.15}{0.92}  & \MErr{56.58}{0.44}    \\
        \bottomrule
    \end{tabular}
    }
\end{table*}

%% file: depd/table-hmean.tex
\begin{table}[!t]
        \centering
        \caption{\label{table:acc}
                Best seen, unseen, and cH-Mean accuracy (\%) of different methods.
                Best results are in bold.}
        \vspace{-1em}
        \adjustbox{width=\columnwidth}{
        \begin{tabular}{lccccccc} \toprule
                                                         & \multicolumn{3}{c}{MIT-States} &  & \multicolumn{3}{c}{UT-Zappos} \\ 
                                                              \cmidrule{2-4}                      \cmidrule{6-8} 
        Model                                            & Seen                & Unseen             & cH-Mean   &  & Seen             & Unseen            & cH-Mean          \\ \midrule
        AttrAsOp~\cite{Nagarajan2018AttributesAO}        & 14.3                & 17.4               & 9.9       &  & 59.8             & 54.2              & 40.8             \\
        RedWine~\cite{Misra2017FromRW}                   & 20.7                & 17.9               & 11.6      &  & 57.3             & \textbf{62.3}     & 41.0             \\
        LabelEmbed+~\cite{Nagarajan2018AttributesAO}     & 15.0                & 20.1               & 10.7      &  & 53.0             & 61.9              & 40.6             \\
        FeatureGen~\cite{Xian2017FeatureGN}              & 24.8                & 13.4               & 11.2      &  & 61.9             & 52.8              & 40.0             \\
        TMN~\cite{Purushwalkam2019TaskDrivenMN}          & 20.2                & 16.1               & 13.0      &  & 58.7             & 60.0              & 45.0             \\
        AFGCL~\cite{Wei2019AdversarialFC}                & \textbf{35.6}       & 17.5               & 14.0      &  & 83.9             & 56.1              & 49.4             \\
        SymNet~\cite{Li2020SymmetryGA}                   & 24.4                & \textbf{25.2}      & 16.1      &  & -                & -                 & -                \\ 
        BMP-Net                                          & 32.9                & 19.3               & \best{16.5}      &  & 83.9      & 60.9              & \best{56.9}      \\ \midrule
        
        AFGCL-960~\cite{Wei2019AdversarialFC}            & 30.8                & 20.6               & 14.9      &  & 84.7             & 58.0              & 50.5             \\
        BMP-Net-960                                      & \textbf{38.6}                & \textbf{21.7}               & \textbf{19.1}      &  & \textbf{87.3}             & \textbf{64.5}              & \textbf{61.1}             \\
        \bottomrule
        \end{tabular}
        }
        \end{table}

%% file: depd/fig-auc.tex
\begin{figure}[t]
    \centering
    \begin{subfigure}[t]{0.47\columnwidth}
        \centering
        \includegraphics[width=1.0\columnwidth, keepaspectratio]{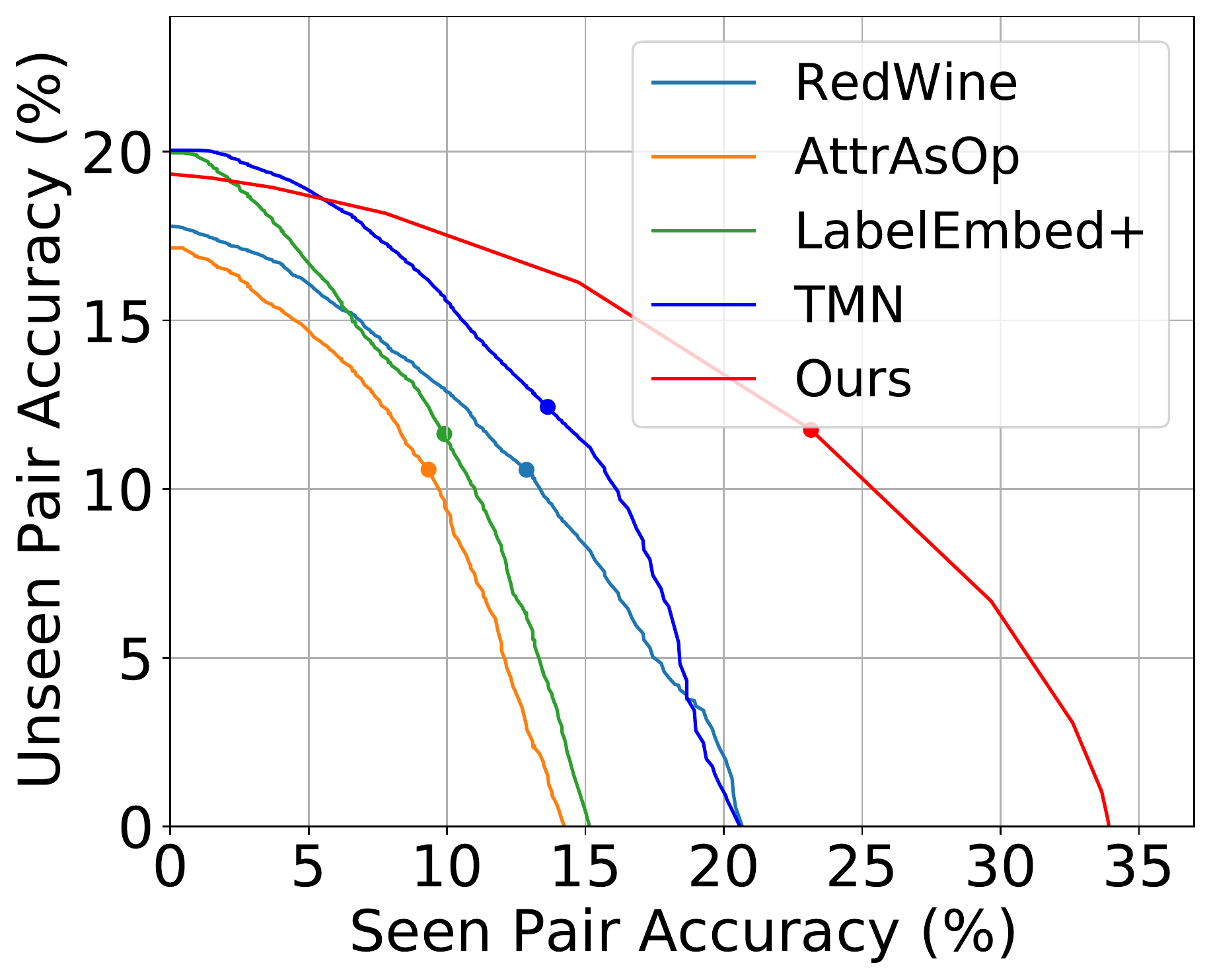}
        \caption{\label{fig:auc} 
        Baselines vs. our model.}
    \end{subfigure}%
    \hfill
    \begin{subfigure}[t]{0.47\columnwidth}
        \centering
        \includegraphics[width=1.0\columnwidth, keepaspectratio]{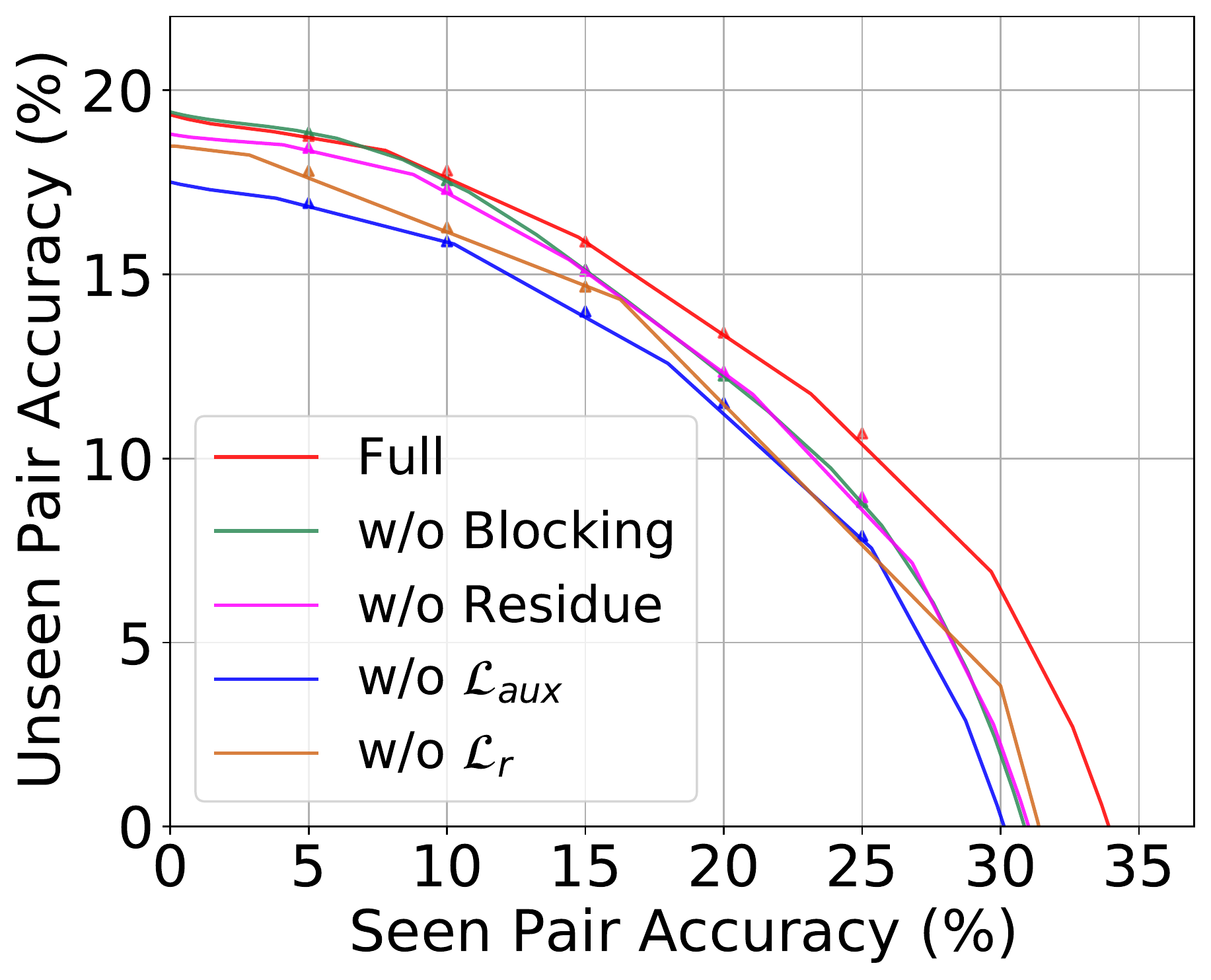}
        \caption{\label{fig:auc_ablation}
        Models for ablation study.}
    \end{subfigure}
	\caption{\label{fig:auc_figs}
        Unseen-seen accuracy on MIT-States under various calibration biases.
        In (a), the circles are the points used to report the cH-Mean in \tabref{table:acc}. 
    }
\end{figure}

%% file: depd/fig-topk.tex
\begin{figure}[t]
    \centering
    \includegraphics[width=1.0\columnwidth, keepaspectratio]{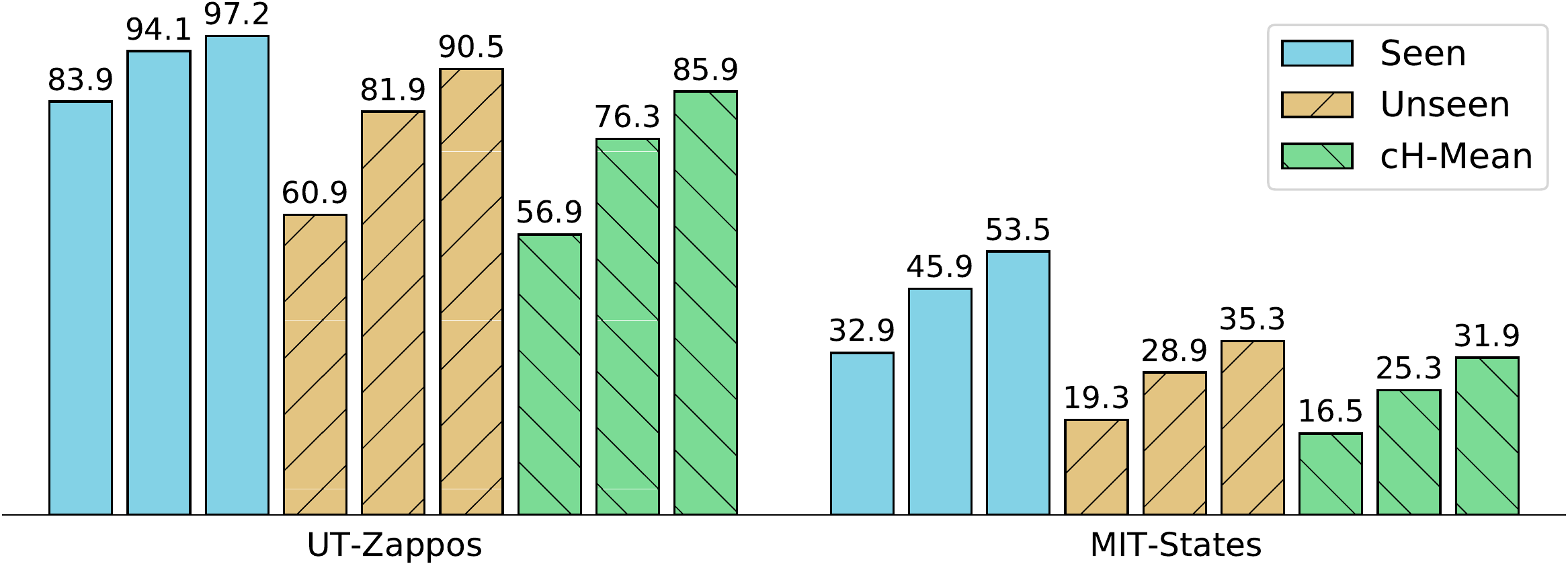}
	\caption{\label{fig:topk}
        Top-1,2,3 accuracy and the corresponding cH-Mean.
        Within each group, from the leftmost to the rightmost are top-1,2,3 accuracy respectively.}
\end{figure}

%% file: depd/table-ablation-lbdr.tex
\begin{table*}[!t]
    \centering
    \caption{\label{table:ablation-lbdr}
        Ablation study with different $\lambda_r$.
        Results in bold are reported as ``full'' in \tabref{table:ablation-study}.}
    \vspace{-1em}
    \adjustbox{width=\textwidth}{
    \begin{tabular}{ll ccccc c ccccc@{}} \toprule
                &  & \multicolumn{5}{c}{MIT-States}                                      & \multicolumn{1}{l}{} & \multicolumn{5}{c}{UT-Zappos}                                     \\ \cmidrule(lr){3-7} \cmidrule(l){9-13}
    $\lambda_r$ &  & Val AUC     & Test AUC    & Seen        & Unseen      & cH-Mean     &                      & Val AUC    & Test AUC   & Seen        & Unseen      & cH-Mean     \\ \midrule
    0           &  & \MErr{5.77}{0.12} & \MErr{3.90}{0.11} & \MErr{30.10}{0.77} & \MErr{18.25}{0.15} & \MErr{14.93}{0.22} &  & \MErr{50.10}{0.71} & \MErr{43.52}{0.92} & \MErr{84.07}{0.34} & \MErr{59.31}{1.09} & \MErr{55.18}{1.03}                     \\
    1           &  & \MErr{6.07}{0.07} & \MErr{4.25}{0.08} & \MErr{31.86}{0.66} & \MErr{18.76}{0.18} & \MErr{15.60}{0.12} &  & \MErr{50.27}{0.45} & \MErr{43.15}{0.87} & \MErr{84.13}{0.20} & \MErr{59.55}{1.01} & \MErr{54.99}{1.63}                     \\
    2           &  & \MErr{6.12}{0.08} & \MErr{4.28}{0.09} & \MErr{31.71}{0.47} & \MErr{18.81}{0.29} & \MErr{15.55}{0.14} &  & \MErr{50.16}{0.44} & \MErr{43.77}{0.86} & \MErr{83.85}{0.21} & \MErr{60.16}{1.08} & \MErr{55.82}{1.30}                     \\
    5           &  & \MErr{\best{6.01}}{0.07} & \MErr{\best{4.31}}{0.09} & \MErr{\best{32.53}}{0.42} & \MErr{\best{18.86}}{0.21} & \MErr{\best{16.01}}{0.18} &  & \MErr{50.43}{0.46} & \MErr{43.68}{0.77} & \MErr{83.81}{0.22} & \MErr{60.18}{0.98} & \MErr{55.31}{1.85}                     \\
    10          &  & \MErr{6.17}{0.07} & \MErr{4.37}{0.09} & \MErr{32.48}{0.39} & \MErr{18.72}{0.24} & \MErr{15.56}{0.13} &  & \MErr{\best{51.12}}{0.47} & \MErr{\best{44.69}}{0.60} & \MErr{\best{83.91}}{0.20} & \MErr{\best{60.15}}{0.92} & \MErr{\best{56.58}}{0.44}                      \\
    20          &  & \MErr{6.15}{0.05} & \MErr{4.41}{0.07} & \MErr{32.44}{0.48} & \MErr{18.92}{0.23} & \MErr{15.79}{0.06} &  & \MErr{50.25}{0.46} & \MErr{43.86}{0.49} & \MErr{84.07}{0.10} & \MErr{60.14}{0.66} & \MErr{56.44}{0.70}                     \\
    50          &  & \MErr{6.06}{0.09} & \MErr{4.30}{0.06} & \MErr{32.06}{0.19} & \MErr{18.73}{0.19} & \MErr{15.60}{0.05} &  & \MErr{50.96}{0.53} & \MErr{44.73}{0.67} & \MErr{83.91}{0.28} & \MErr{60.18}{0.86} & \MErr{56.57}{0.45}                     \\ \bottomrule
    \end{tabular}
    }
\end{table*}

%% file: depd/fig-feat-vis.tex
\begin{figure*}[!t]
	\centering
	\includegraphics[width=1.0\textwidth]{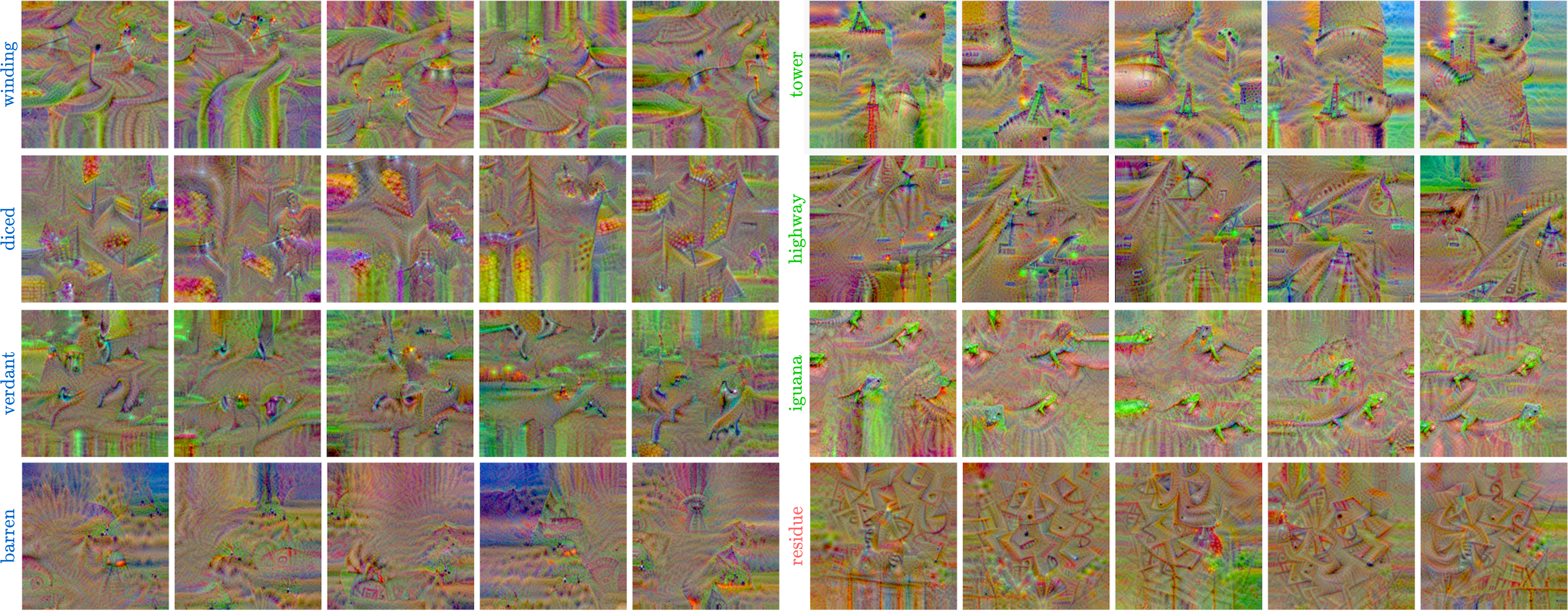}
    \vspace{-2em}
	\caption{\label{fig:feat-vis}
		Examples of visual correspondences.
		The left examples are the visual correspondences of attribute concepts.
		For the right examples, the upper three rows are the visual correspondences of object concepts and the bottom row are those from the residue features.
        Results of five random initializations are shown.
        The vertical texts marks the names of the concepts, where the attributes are in {\color{blue} blue}, the objects are in {\color{green} green}, and the residue is in {\color{red} red}.
        The figure has been enhanced for higher contrast and is best viewed in color/on screen.
	}
\end{figure*}

%% file: depd/fig-failure-analysis.tex
\begin{figure*}[!t]
	\centering
	\includegraphics[width=1.0\textwidth, keepaspectratio]{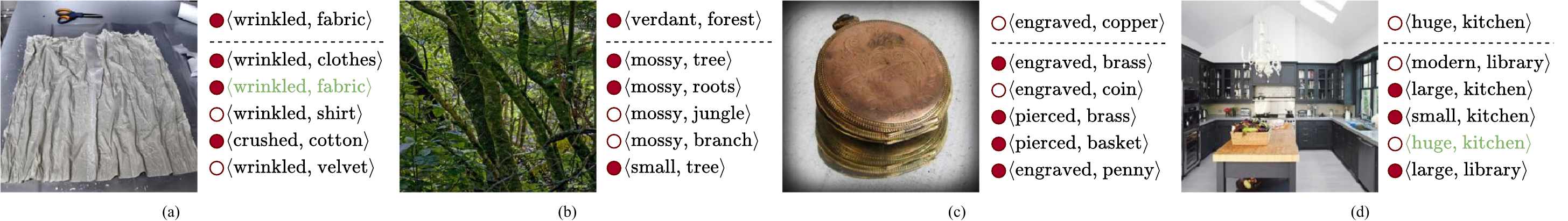}
    \vspace{-2em}
	\caption{ \label{fig:failure-analysis}
    	Examples of failure cases. 
    	For each group, on the left is the input image. 
    	The top row on the right is the ground-truth, and the rows below are top-1 to top-5 predictions.
    	Seen concepts are marked by filled circles.
    	Correct predictions are in \textcolor{green}{green}.
		}
\end{figure*}

%% file: depd/fig-retrieval.tex
\begin{figure*}[!t]
	\centering
	\includegraphics[width=1.0\textwidth, keepaspectratio]{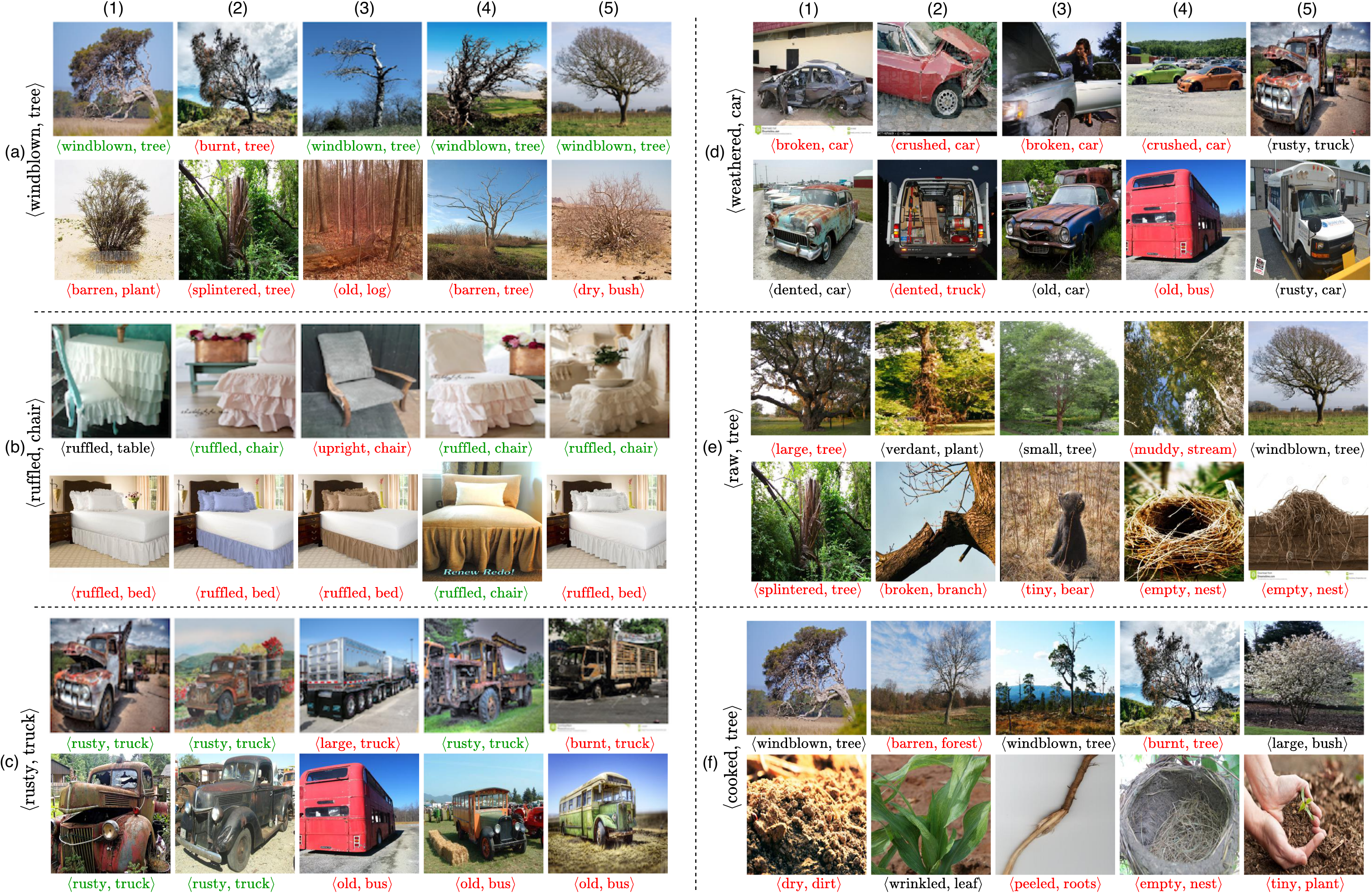}
    \vspace{-2em}
	\caption{ \label{fig:retrieval}
		Retrieval results for composite concepts on all images in MIT-States.
		Column numbered (k) shows the $k^{\text{th}}$ retrieval result, ranked by similarity score.
		(a) through (c) show the result for unseen composite concepts.
		(d) through (f) show the result for ``out-of-the-world'' concepts.
		In each group, the first row shows the result of our model, the second row shows the result of TMN~\cite{Purushwalkam2019TaskDrivenMN}.
		The vertical text shows the query concept. 
		The text under each image is its ground-truth label.
		For all groups, seen concepts are in {\color{red} red}.
		For group (a)-(c), the correct retrievals are in {\color{green} green}.
		The figure is best viewed in color/on screen.
		}
\end{figure*}

%% file: sec_conclusion.tex
\section{Conclusion}
\label{sec:conclusion}

In this work, we address the problem of CZSL on attribute-object pairs.
We propose the BMP-Net that is driven by a key-query based attention mechanism to capture the relations between primitive concepts.
Moreover, we propose a novel blocked message passing mechanism to rectify the bias towards seen pairs and the entanglement between attributes and objects.
Through extensive experiments, we discover several potential extensions to our proposed approach. 
One possible future research direction is to condition the attribute features based on object groups explicitly.
For example, an attribute could have multiple concept features. 
The selection of the specific feature is dependent on the connection between the attribute-object pair provided in an external knowledge base.
It can also help attribute-based domain and task transfer~\cite{Han2012CorrelatedAttributeT,Li2020GradmixMS}.
Another direction is to extend our model to recognize more complex composite concepts whose cardinalities are larger than two (\eg~a \mbox{$\langle \mathsf{white}, \mathsf{wet}, \mathsf{dog} \rangle$} or \mbox{$\langle \mathsf{dog}, \mathsf{on}, \mathsf{road} \rangle$}), where our framework can serve to resolve the challenges due to the sparse or imbalanced data distribution.

%% file: sec_biography.tex
\begin{IEEEbiography}[{\includegraphics[width=1in,height=1.25in,clip,keepaspectratio]{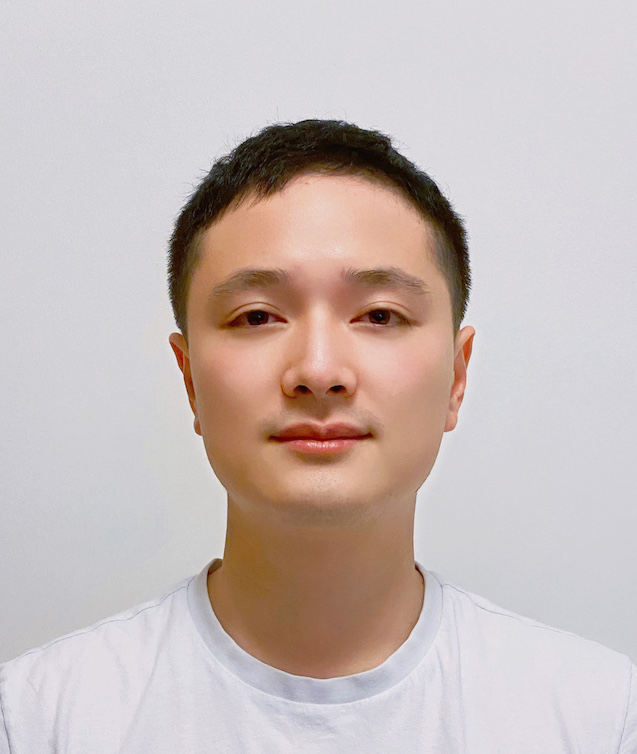}}]{Ziwei~Xu}
    is a PhD candidate at the School of Computing, National University of Singapore.
    He obtained his BEng from the University of Science and Technology of China.
    His current research interests are compositional image/video analysis, knowledge-based visual analysis, and action recognition.
    He is a graduate student member of IEEE since 2017.
\end{IEEEbiography}

\begin{IEEEbiography}[{\includegraphics[width=1in,height=1.25in,clip,keepaspectratio]{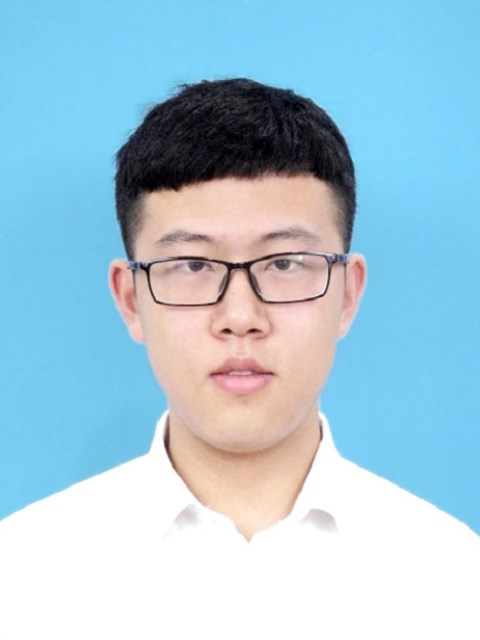}}]{Guangzhi~Wang}
is a PhD candidate at the School of Computing and Institute of Data Science, National University of Singapore.
He obtained his BEng from Zhejiang University.
His current research interests include vision and language, multi-domain learning and knowledge graph.
\end{IEEEbiography}

\begin{IEEEbiography}[{\includegraphics[width=1in,height=1.25in,clip,keepaspectratio]{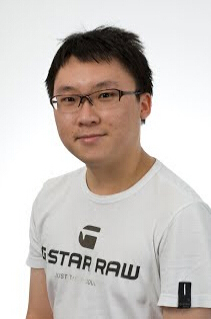}}]{Yongkang~Wong}
	is a senior research fellow at the School of Computing, National University of Singapore. 
	He is also the Assistant Director of the NUS Centre for Research in Privacy Technologies (N-CRiPT). 
	He obtained his BEng from the University of Adelaide and PhD from the University of Queensland. 
	He has worked as a graduate researcher at NICTA's Queensland laboratory, Brisbane, OLD, Australia, from 2008 to 2012. 
	His current research interests are in the areas of Image/Video Processing, Machine Learning, Action Recognition, and Human Centric Analysis. 
	He is a member of the IEEE since 2009.
\end{IEEEbiography}

\begin{IEEEbiography}[{\includegraphics[width=1in,height=1.25in,clip,keepaspectratio]{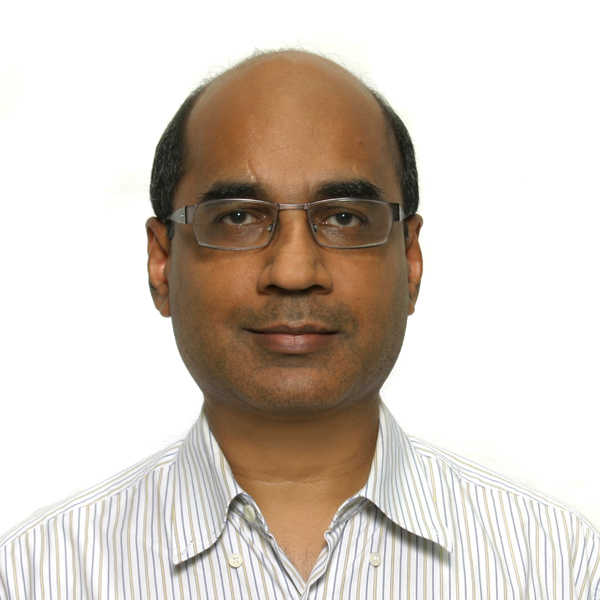}}]{Mohan~S.~Kankanhalli}
	is the Provost's Chair Professor at the Department of Computer Science of the National University of Singapore. 
	He is the director with the N-CRiPT and also the Dean, School of Computing at NUS. 
	Mohan obtained his BTech from IIT Kharagpur and MS \& PhD from the Rensselaer Polytechnic Institute. 
	His current research interests are in Multimedia Computing, Multimedia Security and Privacy, Image/Video Processing and Social Media Analysis. 
	He is on the editorial boards of several journals. 
	Mohan is a Fellow of IEEE.
\end{IEEEbiography}